\documentclass[11pt]{article}
\usepackage[a4paper,margin=2.4cm]{geometry}
\usepackage{mathptmx}
\usepackage[expansion=false]{microtype}
\usepackage{amsmath}
\usepackage{amssymb}
\usepackage{booktabs}
\usepackage{array}
\usepackage{float}
\usepackage{placeins}
\usepackage{textcomp}
\usepackage{graphicx}
\usepackage{url}
\usepackage[numbers,sort&compress]{natbib}
\usepackage{caption}
\usepackage{authblk}

\usepackage[colorlinks=true,linkcolor=blue,citecolor=blue,urlcolor=blue]{hyperref}

\newcommand{\mainfig}[1]{\includegraphics[width=\linewidth,height=0.70\textheight,keepaspectratio]{#1}}
\newcommand{\fullfig}[1]{\includegraphics[width=\textwidth,height=0.92\textheight,keepaspectratio]{#1}}

\graphicspath{{figures/}{./}}
\DeclareGraphicsExtensions{.pdf,.jpg,.jpeg,.png}

\title{\textbf{OPERA: Operator--Residual Feedback for Reliable Autonomous Optical Experiments with Language-Model Agents}}

\author[1,$\dagger$]{Ning Xu}
\author[2,$\dagger$]{Xiang Zheng}
\author[3]{Fuqiang Zhong}
\author[2,*]{Huadong Wang}
\author[4]{Xiaolong Wu}
\author[1,*]{Zhiyuan Liu}
\author[4,*]{Hui Ning}
\affil[1]{Department of Computer Science and Technology, Tsinghua University, Beijing, China}
\affil[2]{ModelBest Inc., Beijing, China}
\affil[3]{School of Artificial Intelligence and Robotics, Hunan University, Changsha, China}
\affil[4]{Northwest Institute of Nuclear Technology, Xi'an, China}
\affil[$\dagger$]{These authors contributed equally to this work.}
\affil[*]{Correspondence: Huadong Wang (\href{mailto:wanghuadong.2018@tsinghua.org.cn}{wanghuadong.2018@tsinghua.org.cn}); Zhiyuan Liu (\href{mailto:liuzy@tsinghua.edu.cn}{liuzy@tsinghua.edu.cn}); and Hui Ning (\href{mailto:ninghuisun@aliyun.com}{ninghuisun@aliyun.com}).}
\date{}

\begin{document}
\maketitle

\begin{abstract}
\noindent Autonomous agents choose actions using scores that may not reflect experimental success. We developed OPERA, an operator--residual framework for optical experiments. It represents experimental actions as optical operators and evaluates their outcomes using physically interpretable residuals. Operators specify executable changes to measurement, control or reconstruction, while residuals report departures from specified physical conditions. The agent uses both to select, combine or generate operators, and physical performance is evaluated independently against a withheld reference. Across three optical tasks, score-only feedback produced score increases without physical improvement in 23.6--39.0\% of decisions, compared with 0.9--1.9\% for operator--residual feedback. Operator--residual feedback increased the probability of reaching and maintaining task targets and reduced experimental budgets. Protocols selected in digital twins were transferred to three optical instruments, and repeated experiments showed a lower projection budget in structured-light reconstruction. Together, operators and residuals guide autonomous decisions using measurable physical evidence.
\end{abstract}

\section{Introduction}\label{sec:intro}

Autonomous agents can design scientific experiments \cite{King2004,King2009} by selecting measurements and updating procedures as data arrive \cite{Burger2020,Szymanski2023}. Related online optimization systems can change experimental settings in response to incoming data \cite{Wigley2016}. Their reliability depends on both the actions they can execute and the feedback used to judge the effects of those actions. Most experimental systems summarize progress with one or a few performance scores. These scores can omit physical conditions that determine validity, allowing measured performance to improve while the experiment itself does not.

Tool-calling language models have expanded the actions available to autonomous systems by invoking external tools \cite{Schick2023Toolformer,Qin2023ToolLLM}. They can also execute multistep plans and revise actions using intermediate feedback \cite{Yao2023ReAct,Shinn2023Reflexion}. In scientific workflows, language-model agents can combine analytical operations and generate executable procedures \cite{Boiko2023Coscientist,Bran2024ChemCrow}. They have also been connected to microscopy and other laboratory instruments \cite{Yang2025AutonomousMicroscopy,Vriza2026Instruments,Chen2026XRay}. These capabilities increase experimental flexibility, but planning alone cannot recover information absent from the feedback. Repeated optimization of an incomplete score can instead exploit what that score omits. This is a form of specification gaming, in which optimization of the measured objective diverges from the intended outcome \cite{Goodhart1975,Manheim2019}. Related studies show that repeated optimization against learned or incomplete rewards can amplify this mismatch \cite{Pan2024ICRH,Gao2023,Coste2023}.

We developed OPERA, a framework for autonomous optical experiments that organizes executable operators and diagnostic residual feedback. An optical operator is a checked executable change to phase shifts, projected patterns, modulation masks, measurement locations or reconstruction procedures. The agent can select an existing operator, combine operators or generate a new procedure through the same interface. After execution, physically interpretable residuals quantify departures from specified physical conditions using the resulting observations. Operators determine what the experiment can do. Residuals report which physical conditions still require correction. The agent uses both to select the next operator. Physical performance is evaluated separately against a withheld reference and never enters the decision loop.

Optical experiments provide a useful test of this framework because their actions are programmable, their observations are high-dimensional and a single performance score may not capture all conditions defining a valid outcome. We implemented OPERA in beam shaping, structured-light three-dimensional reconstruction and interferometry. Together, these tasks cover the control of light distributions, three-dimensional measurement and phase reconstruction. In each task, residuals were calculated from observations available during execution, without access to the withheld physical reference.

We evaluated the OPERA framework across the three optical tasks using four tool-calling language models under fixed execution budgets. The decision experiments were conducted in digital twins, and selected protocols were subsequently transferred to physical instruments. Matched comparisons held the operator libraries and execution budgets constant while changing the feedback available to the agent. Operator--residual feedback reduced score increases without physical improvement, increased the probability of reaching and maintaining task targets and reduced experimental budgets relative to the four primary reference strategies. Comparisons with stronger algorithmic baselines defined the conditions under which the budget advantage remained. Separate analyses examined how residual information affected operator selection, independently validated generated strategies and evaluated transfer to optical hardware.
\medskip
\noindent Our main contributions are as follows.
\begin{itemize}
\item \textbf{Framework.} OPERA represents experimental actions as typed, checked optical operators and returns physically interpretable residual feedback through a common interface. Diagnostic residuals ($\mathrm{L}_{1}$), the visible performance score ($\mathrm{L}_{2}$) and offline physical performance against a withheld reference ($\mathrm{L}_{3}$) are kept strictly separate, so score increases without physical improvement become measurable.
\item \textbf{Feedback determines physical validity.} In a matched factorial study across three optical tasks and four tool-calling language models (7,486 valid decisions over 90 independent problems), score-only feedback produced score increases without physical improvement in 23.6--39.0\% of decisions, versus 0.9--1.9\% under operator--residual feedback. The separation persisted across all 25 tolerance combinations and with model reasoning disabled or enabled.
\item \textbf{Target attainment and budget.} Across 270 independent test problems, operator--residual feedback reached and maintained task targets in 75.8\% of cases, against 33.9--60.1\% for four matched reference strategies, while using 61.7\% of the available budget against 70.8--86.0\%. Comparisons with constrained Bayesian optimization, task-specific methods and a fixed residual rule located the remaining advantage under distribution shift.
\item \textbf{Mechanism.} Controls that vary the representation of the same residual information (aggregate scalar, anonymous, labelled and described channels) quantify the value of channel structure and physical semantics, and a counterfactual operator audit shows that residual feedback increases selection of the operator with the largest offline physical improvement, from 2\% to 34\% in structured-light reconstruction.
\item \textbf{Generation and hardware transfer.} The same interface supports strategy generation with independent physical validation and prior-art assessment (53 of 60 prespecified attempts executable; all planted failure controls rejected), and protocols selected in digital twins transfer to three optical instruments, with a repeated projection-budget advantage in structured-light reconstruction.
\end{itemize}

\section{Related work}\label{sec:related}

\subsection{Autonomous experimentation}
Closed-loop systems have automated hypothesis generation and experiment selection in functional genomics \cite{King2004,King2009}, mobile robotic experimentation in chemistry \cite{Burger2020} and autonomous solid-state synthesis of inorganic materials \cite{Szymanski2023}, and online optimization has tuned ultracold-atom experiments directly from incoming data \cite{Wigley2016}. Language-model agents extend this line of work. They have combined analytical operations and generated executable procedures in chemistry \cite{Boiko2023Coscientist,Bran2024ChemCrow}, and they have been connected to microscopes, beamlines and other laboratory instruments \cite{Yang2025AutonomousMicroscopy,Vriza2026Instruments,Chen2026XRay}, with related closed-loop control demonstrated in optical trapping \cite{Selin2026SmartTrap}. These systems typically summarize progress with one or a few performance scores. OPERA structures the feedback itself, returning residuals that report the physical conditions a scalar score can omit, and evaluates physical success against a reference withheld from the decision loop.

\subsection{Tool use and agentic planning with language models}
Language models can learn to invoke external tools \cite{Schick2023Toolformer} and to operate large tool collections \cite{Qin2023ToolLLM}. ReAct interleaves reasoning with actions \cite{Yao2023ReAct}, and Reflexion revises behaviour using verbal feedback on intermediate outcomes \cite{Shinn2023Reflexion}. These mechanisms expand what an agent can execute and how it can replan, but they operate on whatever feedback the environment returns. OPERA is complementary. It specifies the executable action space through typed optical operators and supplies the diagnostic residual feedback on which such replanning can act.

\subsection{Specification gaming and reward hacking}
Optimizing a measured objective can diverge from the intended outcome, a failure summarized by Goodhart's law \cite{Goodhart1975,Manheim2019}. Optimization against learned or incomplete reward models degrades true performance as optimization pressure increases \cite{Gao2023}, reward-model ensembles only partially mitigate the effect \cite{Coste2023}, and feedback loops can drive in-context reward hacking in language models \cite{Pan2024ICRH}. Preference learning \cite{Christiano2017} and inverse reward design \cite{HadfieldMenell2017} treat the specification itself as uncertain, and analytical redundancy uses multiple physical checks to expose faults hidden from a single output \cite{Chow1984}. OPERA carries this concern into physical experimentation. It defines score increases without physical improvement against a withheld reference and measures how structured residual feedback changes their rate.

\subsection{Safe action selection and experimental design}
Bayesian optimization provides sample-efficient search over expensive experiments \cite{Jones1998,Snoek2012}. Physical requirements can be incorporated during action selection \cite{Gardner2014,Sui2015}. Model-based rules can restrict execution \cite{Hewing2020,Wabersich2021}, while experimental-design methods guide measurement selection \cite{Foster2019,Rainforth2023}. OPERA addresses a complementary problem by representing executable actions and diagnostic residuals through a common interface. Its action-selection policy remains replaceable, allowing these approaches to be used within the same framework.

\section{The OPERA framework}\label{sec:framework}

\subsection{Framework overview}

\begin{figure}[p]
\centering
\includegraphics[width=\textwidth,height=\dimexpr\textheight-52pt\relax,keepaspectratio]{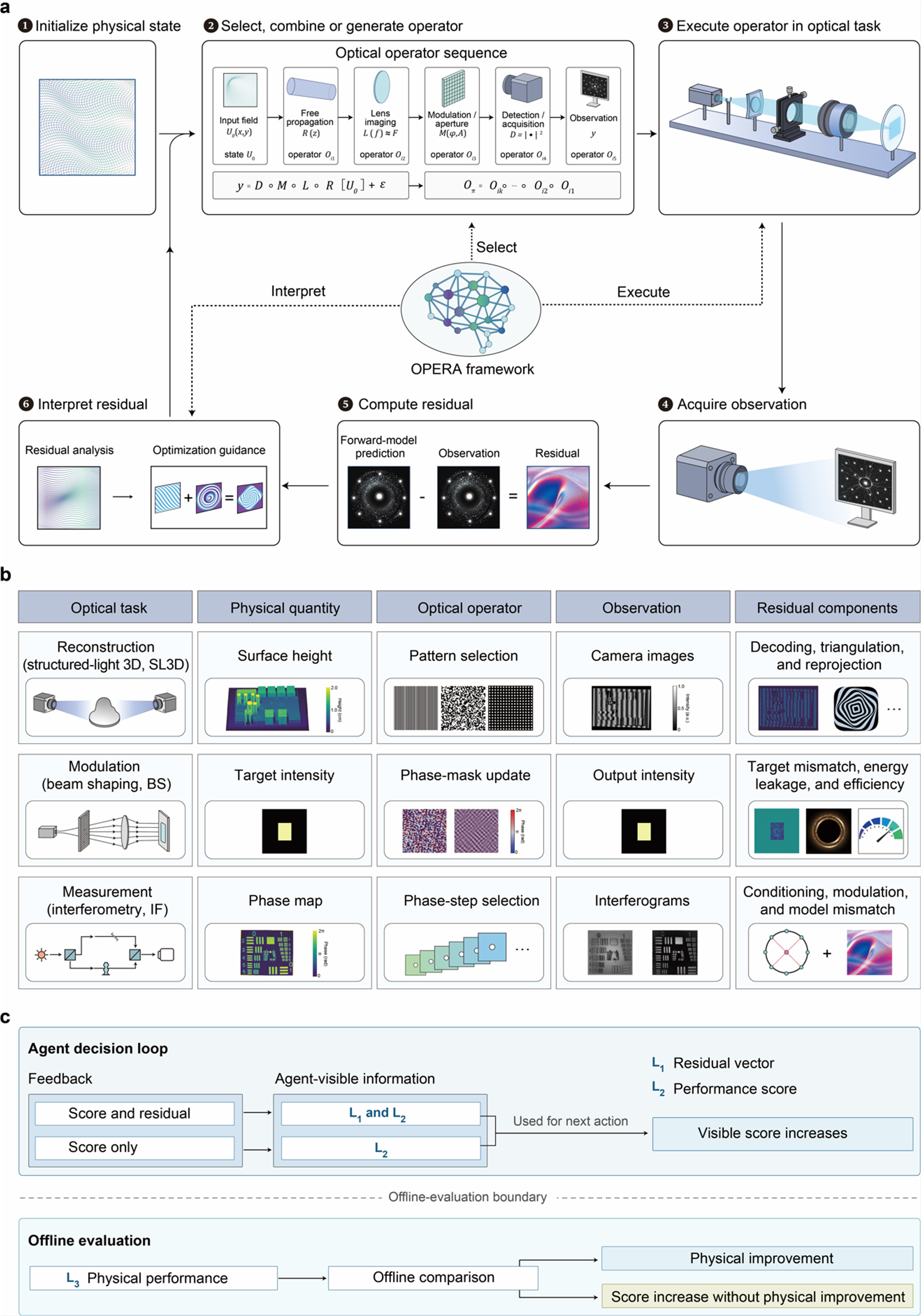}
\caption{\textbf{Operator--residual feedback in optical experiments.}}
\label{fig:framework}
\end{figure}

The OPERA framework represents an optical experiment as a sequence of fixed physical transformations and selectable operators. The fixed transformations include the Fourier transform produced by a lens and the convolution describing free-space propagation. At each step, the agent can select a cataloged operator, combine existing operators or generate an executable procedure from permitted operations. Each proposed action is checked before execution and charged to the same task-specific resource budget (Fig.~\ref{fig:framework}a and Supplementary Tables~1 and 2).

Execution returns an observation and the feedback used for the following decision. Under operator--residual feedback, the agent receives a structured residual vector, $\mathrm{L}_{1}$, together with a scalar performance score, $\mathrm{L}_{2}$. Score-only feedback supplies $\mathrm{L}_{2}$ alone. Each component of $\mathrm{L}_{1}$ measures departure from a specified physical condition that can be evaluated from the observations. Physical performance, $\mathrm{L}_{3}$, is calculated offline against a withheld reference and is never available to the agent (Fig.~\ref{fig:framework}c). The three quantities therefore distinguish diagnostic feedback, the score used during execution and the independent evaluation of physical success.

\begin{figure}[!t]
\noindent{\small \textbf{\figurename~\thefigure\ (continued).} \textbf{a}, The OPERA framework represents an optical task as a sequence of fixed transformations and selectable optical operators. The language-model agent selects, combines or generates an operator through a common executable interface. The operator is checked, executed and charged to the task-specific resource budget. The resulting observation is then used to calculate feedback for the next decision. \textbf{b}, Implementations in structured-light three-dimensional reconstruction (SL3D), beam shaping (BS) and interferometry (IF). Each row links the physical quantity of interest to the available operators, resulting observations and residual components. \textbf{c}, Information available during execution and offline evaluation. Score + residual supplies the residual vector $\mathrm{L}_{1}$ and performance score $\mathrm{L}_{2}$, whereas Score only supplies $\mathrm{L}_{2}$. Each component of $\mathrm{L}_{1}$ measures departure from a specified observable physical condition. Offline evaluation compares the submitted result with a withheld reference to calculate the task-specific physical performance $\mathrm{L}_{3}$. Neither the reference nor $\mathrm{L}_{3}$ enters the decision loop. Changes in $\mathrm{L}_{2}$ and $\mathrm{L}_{3}$ are compared offline to distinguish physical improvement from a score increase without physical improvement.}
\end{figure}
\FloatBarrier

\subsection{Decision process and operator interface}

The OPERA framework treats an experiment as a partially observed decision process. A hidden state $x$ contains the task-specific physical reference, such as a surface-height field, phase map or target intensity distribution. This state is withheld during execution and used only for offline evaluation. At decision step $t$, the information available to the language model is

\begin{equation}
s_t = \left(T, \mathcal{O}_t, b_t, \{(O_\tau, y_\tau, \mathrm{L}_{2,\tau}, R_\tau)\}_{\tau \le t}\right),
\end{equation}

where $T$ is the task specification, $\mathcal{O}_t$ is the current operator library, $b_t$ is the remaining resource budget and the final term records the operators, observations, scores and residuals obtained so far. Under score-only feedback, $R_\tau$ is omitted from the information returned to the model.

Operators belong to four functional classes: MEASURE, MODULATE, RECONSTRUCT and ANALYZE. MEASURE and MODULATE consume task-specific physical resources such as detector frames, projected patterns or phase-mask evaluations. RECONSTRUCT and ANALYZE act on recorded data and have zero physical cost. Sequentially composed operators have additive cost,

\begin{equation}
O = O^{(m)} \circ \dots \circ O^{(1)}, \qquad c(O) = \sum_{j=1}^m c\left(O^{(j)}\right).
\end{equation}

The language model selects a cataloged operator, combines existing operators or generates a procedure from the permitted primitives. Generated procedures call the same typed application programming interface and enter the library only after validation. An external harness serializes the current state, checks each call and records resource use, so every model call receives the same explicit execution history rather than relying on model-side memory.

The scalar score $\mathrm{L}_{2}$ and the structured residual vector $\mathrm{L}_{1}$ are both calculated from observations available during execution. Each component of $\mathrm{L}_{1}$ tests a specified observable physical condition. The task-specific physical performance $\mathrm{L}_{3}(g,x)$ is calculated only after submission by comparing the final design or reconstruction $g$ with the withheld state $x$. Neither $x$ nor $\mathrm{L}_{3}$ is available to the decision loop.

\begin{center}
\fbox{\begin{minipage}{\dimexpr\linewidth-2\fboxsep-2\fboxrule\relax}
\small
\textbf{Algorithm 1 | Operator--residual decision loop}\par\smallskip
\textbf{Inputs:} task specification $T$; initial operator library $\mathcal{O}_0$; componentwise resource limits $B$.\par
\begin{enumerate}
\item Initialize the execution history $H_0=\varnothing$ and resource expenditure at zero.
\item While a valid operator can be executed within $B$:
  \begin{enumerate}
  \item Construct $s_t$ from $T$, $\mathcal{O}_t$, the remaining resources and $H_t$.
  \item Let $\pi_\theta(s_t)$ select, compose or generate the next operator $O_{t+1}$ through the typed application programming interface. If a newly generated operator passes the platform checks, append it to $\mathcal{O}_t$.
  \item Execute $O_{t+1}$, acquire observation $y_{t+1}$ and charge its componentwise cost $c(O_{t+1})$.
  \item Compute the visible score $\mathrm{L}_{2,t+1}$ and the observable residual vector $R_{t+1}=\mathrm{L}_{1,t+1}$ from the accumulated observations.
  \item Return $\mathrm{L}_{2,t+1}$ and $R_{t+1}$ under operator--residual feedback, or $\mathrm{L}_{2,t+1}$ alone under score-only feedback, and append the execution record to $H_t$.
  \end{enumerate}
\item Submit the final reconstruction or design $g$. Compute $\mathrm{L}_{3}(g,x)$ only during offline evaluation; neither the withheld state $x$ nor $\mathrm{L}_{3}$ is read by the decision loop above.
\end{enumerate}
\end{minipage}}
\end{center}

Before simulator execution, the command interface also checked the optical layout. Components were represented as nodes in a directed graph and propagation paths as typed edges. A proposed operator sequence was accepted only if it retained a continuous path from source to sample and detector.

\subsection{Task instantiations of operators and residuals}

We implemented these operator and residual roles in beam shaping, structured-light three-dimensional reconstruction and interferometry. Operators determine how the observation can be changed, whereas residuals indicate which physical conditions require correction. In beam shaping, operators changed phase masks, and residuals detected target mismatch or light outside the intended region. In structured-light reconstruction, operators selected projected patterns and reconstruction steps, while residuals identified unusable depth measurements and incomplete coverage. In interferometry, operators selected phase shifts and reconstruction procedures, while residuals tested how well the recorded fringes agreed with the measurement model (Fig.~\ref{fig:framework}b and Supplementary Table~3).

\section{Experimental setup}\label{sec:setup}

This section specifies the tasks, models, reference strategies and endpoints used in the evaluation. Full prespecified definitions, operator sets and residual channels appear in the Supplementary Information provided as ancillary files.

\subsection{Optical tasks and problem sets}

The three digital twins represented beam shaping, structured-light three-dimensional reconstruction and interferometry. Beam-shaping operators followed iterative Fourier-domain and mixed-region phase-design methods \cite{Gerchberg1972,Pasienski2008}. Structured-light operators used Gray-code and phase-shifting procedures \cite{Salvi2004,ZhangHuang2006}. Interferometric reconstruction followed established phase-shifting estimation methods \cite{Creath1988,Surrel1996}. The operator sets, distribution shifts, budget caps, targets and tolerances are listed in Supplementary Table~2; all residual channels and their observable checks are listed in Supplementary Table~3.

The independent problem was the unit of statistical inference. Each problem fixed the target, forward model, apparatus parameters, noise distribution, calibration state and withheld reference. A familiar condition and a prespecified shifted condition were generated from the same problem root. Model, seed, prompt and feedback conditions were nested within that root. The study used 90 development problems and 270 independent test problems, divided equally among the three tasks. Development problems were used to set residual scales, baseline hyperparameters and difficulty covariates; they were not included in test estimates. The full design and failure-handling rules are summarized in Supplementary Table~1.

Problem difficulty was calculated before agent action from prespecified task-specific observables. Full definitions and transformations are provided in Supplementary Methods.

\subsection{Language models and execution harness}

The language models were \texttt{gpt-5.1-2025-11-13}, \texttt{deepseek-v4-pro}, \texttt{qwen3.7-max-2026-06-08} and \texttt{claude-opus-4-7}. DeepSeek v4 pro was accessed in April 2026. Models entered the same harness through their application programming interfaces, and each task prompt was loaded as a fixed skill module. The harness, operator schema and execution checks were held constant across models and experimental arms except for the prespecified feedback, instruction and reasoning-mode changes. Calls used a temperature of 0 with no \texttt{top\_p} parameter. Reasoning modes were disabled in the main experiments. Five numerical seeds controlled independent simulator data streams and were separate from language-model sampling.

\subsection{Feedback conditions and outcome definitions}

The matched two-by-two experiment crossed feedback with instruction. Operator--residual feedback returned $\mathrm{L}_{1}$ and $\mathrm{L}_{2}$; score-only feedback returned $\mathrm{L}_{2}$ alone. The Neutral instruction requested the next valid operator within the remaining budget. The Score-priority instruction instead prioritized an immediate increase in $\mathrm{L}_{2}$. Task states, operator libraries and budgets were unchanged across the four cells.

A decision was classified as a score increase without physical improvement when the increase in $\mathrm{L}_{2}$ exceeded its task-specific test--retest tolerance and the offline change in $\mathrm{L}_{3}$ did not exceed its corresponding tolerance. The analysis contained 7,560 scheduled decisions, of which 7,486 were valid and 74 were invalid or failed. Invalid and failed attempts remained in the complete accounting reported in Supplementary Table~4. Claude Opus 4.7 contributed a separate subset of 60 problems and 411 valid decisions and was not included in the paired factorial contrasts across models.

Robustness was assessed in two prespecified analyses. The first recalculated event rates for all 25 combinations formed by scaling the $\mathrm{L}_{2}$ and $\mathrm{L}_{3}$ tolerances (Supplementary Table~5). The second crossed feedback with reasoning disabled or enabled for GPT 5.1 while holding the problem roots, prompts, operators, budgets and five seeds fixed (Supplementary Table~6).

\subsection{Reference strategies, algorithmic baselines and budget endpoints}

The four primary reference strategies were a fixed expert sequence, random acquisition from the valid operator set, metric-guided selection using $\mathrm{L}_{2}$ alone and vector-matched feedback. The vector-matched arm received synthetic channels matched to the residual vector in dimension, update frequency and numerical distribution but unrelated to the current physical state. Additional comparisons used constrained Bayesian optimization \cite{Jones1998,Snoek2012}, task-specific optical methods and a fixed residual rule (Supplementary Tables~15--17). The fixed rule received the same residual channels and candidate operators as the language model but selected actions with a prespecified task-specific ranking. The six executable strategies used to calculate Best policy at each budget are listed in Supplementary Table~14. That curve is a pointwise visual reference and is not itself an executable strategy.

First-crossing budget was the earliest expenditure at which offline $\mathrm{L}_{3}$ reached the task target. The primary endpoint required $\mathrm{L}_{3}$ to remain above the target through the end of evaluation. Its normalized budget was

\begin{equation}
b_a = \frac{B_{\mathrm{dur}}}{B_{\max}},
\end{equation}

where $B_{\mathrm{dur}}$ is the earliest expenditure after which the target remained satisfied and $B_{\max}$ is the task-specific cap. A trajectory that never reached the target, or later fell below it, was assigned the cap. Display-cohort summaries, problem-level confirmatory endpoints, maintenance-window analyses and full algorithmic-baseline outcomes are reported in Supplementary Tables~7--17.

A complementary analysis reported resource use and the probability of reaching and maintaining the target separately. It used the same 54,000 trajectory records, retained trajectories through the budget cap and averaged repeated model and seed records within each of 270 problem roots before calculating paired contrasts (Supplementary Table~12).

\subsection{Residual-representation controls and operator audits}

Three controls separated the amount of residual information from its interpretation. The anonymous-vector control returned the true residual values after removing channel identities and permuting their positions. The labelled-vector control restored stable names, units, directions, thresholds and update times but omitted the descriptions of the associated physical conditions. The composite-scalar control replaced the residual vector with one equally weighted aggregate violation score. Its construction and paired analysis are reported in Supplementary Methods and Supplementary Table~18.

The offline operator audit cloned the simulator state at a recorded decision and evaluated the selected operator together with five alternatives over the same horizon. Each branch was discarded after its residual change and offline $\Delta\mathrm{L}_{3}$ were recorded; branches did not alter the live trajectory or consume its budget. The audit measured whether the agent selected the candidate with the largest physical improvement (Supplementary Algorithm~2, Appendix Fig.~\ref{edfig:choice} and Supplementary Table~19). Separate controlled stress tests examined residual omission, monitoring and injected faults outside the primary three-task evaluation (Appendix Fig.~\ref{edfig:controlled_stress}).

A task-specific multi-output ridge model was fitted on development transitions to predict the early change in each residual from the visible state and operator configuration. It did not receive $\mathrm{L}_{3}$, model identity or condition labels. Residual prediction accuracy was calculated from the first three transitions of seed 1. Downstream budget reduction was then calculated from separate trajectories using seeds 2--5, preventing the same trajectory from contributing to both quantities. Regression definitions and held-out results are reported in Supplementary Methods and Supplementary Tables~20 and 21.

\subsection{Strategy generation and independent validation}

The generation study contained 60 prespecified attempts: 12 from score-only feedback, 24 from operator--residual feedback and 24 from a dedicated synthesis condition. The latter used a Neutral instruction with residual feedback to generate executable routines without suggesting a candidate solution. These groups yielded 11, 21 and 21 executable records, respectively; all seven non-executable attempts remained in the generation denominators (Supplementary Table~22).

Each executable strategy was frozen and evaluated on 12 independent test problems without further language-model calls. Validation used task-specific offline statistics because the three physical outputs were not directly comparable. Strategy generation, physical validation and prior-art assessment were evaluated separately. Detailed strategy outcomes, validation definitions and prior-art assignments are reported in Supplementary Tables~23--25. Any $\mathrm{L}_{1}$ pass/fail entry in the strategy table was calculated after generation and does not imply that a score-only generation received residual feedback. Secondary checks of interface compliance and optical reasoning were conducted separately from the primary experiments (Supplementary Tables~26 and 27).

\subsection{Hardware transfer protocol}

Protocols were frozen before transfer, and hardware residuals were not returned to the agent. The structured-light system used a Daheng Imaging CMOS camera and an Anhua Optoelectronics projector. Three configurations were evaluated with six paired technical repeats per configuration. An independent dense reference was acquired with a calibrated VK-X3000 laser-confocal surface profiler (Keyence) at a lateral sampling interval of 2.5~\textmu m and a stated axial accuracy of 0.5~\textmu m. Five repeated reference scans yielded an axial standard deviation of 1.2~\textmu m. Reference and structured-light reconstructions were registered with the fiducial plate and compared over the same prespecified support mask. The dense reference was unavailable during protocol selection and hardware execution.

The beam-shaping apparatus used a 632.8-nm He--Ne laser and a reflective PLUTO-2.1 liquid-crystal-on-silicon spatial light modulator (HOLOEYE; $1920 \times 1080$ pixels; 8.0~\textmu m pitch). The modulator displayed the frozen phase mask, and a Fourier lens formed the field on the detector. Spot width is reported in Airy units, which normalize the measured width to the diffraction-limited Airy scale of the apparatus.

The interferometry apparatus used an LH3000 dual-frequency laser interferometer system (Leice Technology; vacuum wavelength, 632.99~nm), a beam splitter, a reflective 1951 USAF target and an Iris 9 detector (Photometrics). The phase-step sequence was transferred directly from the digital twin. Because phase is inferred from the acquired intensity sequence rather than measured directly by the detector, hardware performance was evaluated from registered intensity-pattern agreement, fringe modulation and normalized absolute error. Complete instrument specifications, reconstruction metrics and hardware target definitions are provided in Supplementary Methods.

\subsection{Statistical analysis}

Repeated conditions, language models, seeds, trajectories and decisions were kept within their independent problem root during aggregation and resampling. Unless stated otherwise, intervals are two-sided 95\% percentile intervals from 5,000 bootstrap resamples of independent problem roots \cite{Efron1979,Efron1994}. Primary paired test-set contrasts used two-sided Monte Carlo sign-flip tests with 50,000 draws. Audits containing three to six problem blocks used exact sign-flip enumeration. Holm adjustment controlled the family-wise error rate within each prespecified family of contrasts. The released trajectory records were also re-executed with the released deterministic backend before aggregate summaries were recomputed; the procedure is given in Supplementary Algorithm~3.

\section{Results}\label{sec:results}

\subsection{Operator--residual feedback aligns scores with physical outcomes}\label{sec:score_physical_improvement}

We tested how feedback affected physical validity by comparing score-only and operator--residual feedback under two instructions. The Neutral instruction asked the agent to select the next valid operator within the remaining budget. The Score-priority instruction explicitly prioritized an immediate increase in $\mathrm{L}_{2}$. The problems, operators and budgets were otherwise unchanged, so the comparison altered the emphasis placed on the visible score.

\begin{figure}[H]
\centering
\mainfig{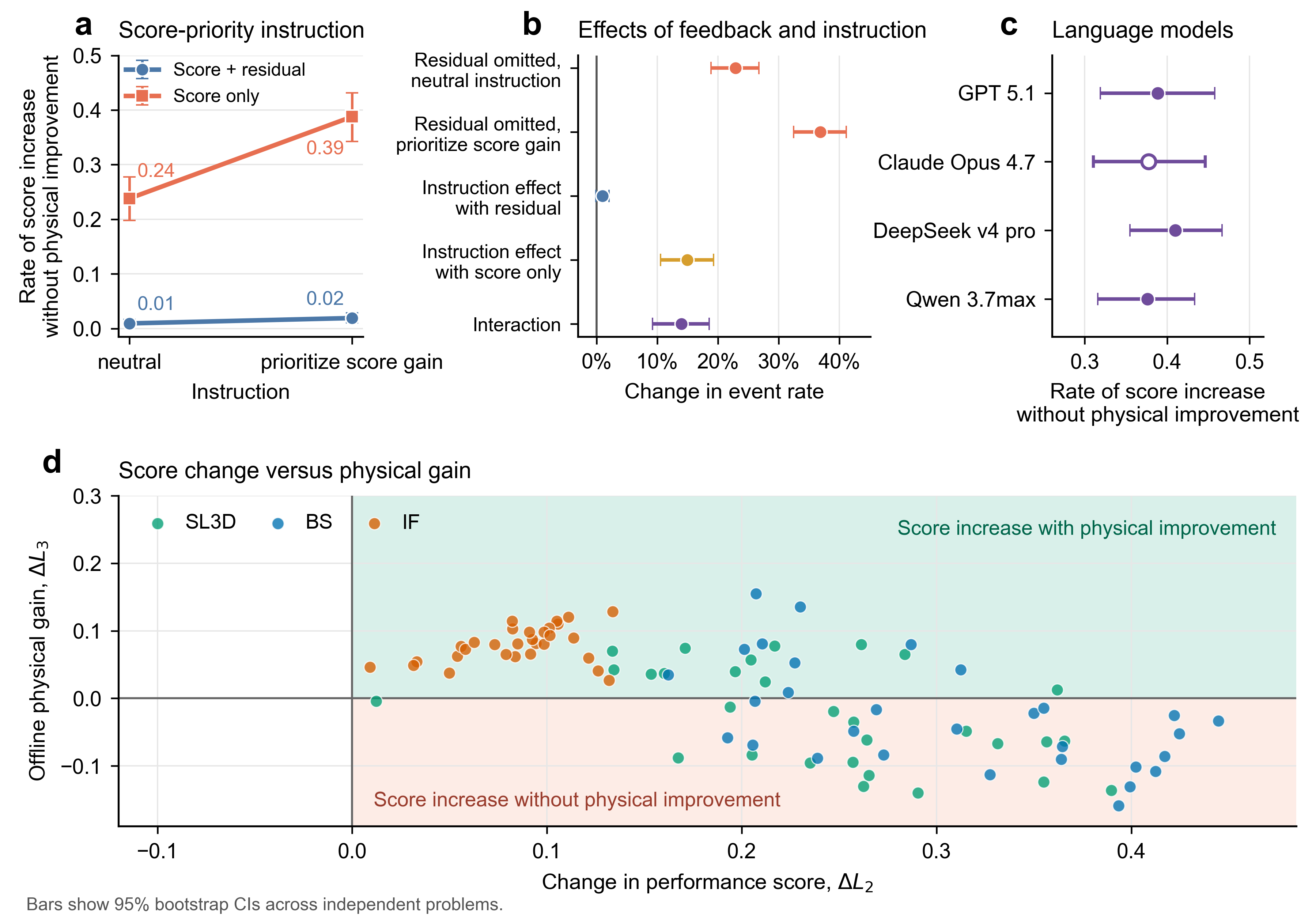}
\caption{\textbf{Operator--residual feedback aligns scores with physical outcomes.}
\textbf{a}, Rate of score increase without physical improvement in the matched factorial experiment crossing score-only or operator--residual feedback with Neutral or Score-priority instruction. An event was recorded when $\mathrm{L}_{2}$ increased beyond its measurement tolerance without improvement in offline $\mathrm{L}_{3}$.
\textbf{b}, Effects of feedback, instruction and their interaction. The interaction is a difference-in-differences; positive values indicate that residual omission increased the event rate more under the Score-priority instruction than under the Neutral instruction.
\textbf{c}, Language-model-specific event rates under score-only feedback and the Score-priority instruction. DeepSeek v4 pro, GPT 5.1 and Qwen 3.7max entered the complete factorial design; Claude Opus 4.7 was evaluated in a separate sensitivity subset. In \textbf{a--c}, points show estimates and bars show two-sided 95\% confidence intervals from 5,000 bootstrap resamples of independent problems.
\textbf{d}, Change in $\mathrm{L}_{2}$ versus change in offline $\mathrm{L}_{3}$ under score-only feedback and the Score-priority instruction. The shaded lower-right region marks score increases without physical improvement. Decisions were first aggregated within language model and then within independent problem. The experiment contained 90 independent problems, with language models and decisions nested within each problem. BS, beam shaping; SL3D, structured-light three-dimensional reconstruction; IF, interferometry.}
\label{fig:score_physical_improvement}
\end{figure}

A decision was counted as a score increase without physical improvement if $\mathrm{L}_{2}$ rose beyond its measurement tolerance while offline $\mathrm{L}_{3}$ did not improve. Under the Neutral instruction, the event rate was 23.6\% with score-only feedback and 0.9\% with operator--residual feedback. Under the Score-priority instruction, the corresponding rates were 39.0\% and 1.9\% (Fig.~\ref{fig:score_physical_improvement}a,b and Supplementary Table~4). Placing greater emphasis on the visible score therefore widened the discrepancy mainly when residual feedback was absent.

In beam shaping, the score could rise as light moved outside the intended region because off-target energy was not fully represented in the scalar score. In structured-light reconstruction, apparent coverage could increase without additional valid depth measurements because coverage alone did not establish valid geometry. Such discrepancies were less frequent in interferometry, where improvements in the intensity fit more often accompanied better phase reconstruction (Fig.~\ref{fig:score_physical_improvement}d). OPERA returned separate residuals for the physical conditions omitted from the scalar score, allowing the agent to distinguish a higher score from physical progress. The separation between feedback conditions persisted across the evaluated language models, all 25 tolerance combinations and a separate analysis with reasoning disabled or enabled (Fig.~\ref{fig:score_physical_improvement}c and Supplementary Tables~5 and 6).

\FloatBarrier

\subsection{Operator--residual feedback improves target attainment and stability}\label{sec:budget}

After establishing closer alignment between the visible score and physical performance, we evaluated target attainment and stability. Success required $\mathrm{L}_{3}$ to reach the task-specific target and remain above it until the end of the evaluation. The four primary reference strategies used a fixed sequence, sampled randomly from the same operator set, selected actions using $\mathrm{L}_{2}$ alone or received a feedback vector matched in size and timing but unrelated to the physical state (Supplementary Tables~7--12).

Across 270 independent test problems, operator--residual feedback reached and maintained the target in 75.8\% of cases, compared with 33.9--60.1\% across the four reference strategies (Supplementary Table~12). Initial target crossing did not always persist. In shifted interferometry, 12 of 40 operator--residual trajectories that reached the target subsequently fell below it, compared with 9 of 11 trajectories under the fixed schedule. Terminal $\mathrm{L}_{3}$ was 0.94 and 0.10, respectively. Operator--residual feedback therefore reduced but did not eliminate later loss of target performance (Appendix Figs.~\ref{edfig:budget_trajectories}--\ref{edfig:confirmatory_budget} and Supplementary Tables~9 and 11).

\begin{figure}[H]
\centering
\mainfig{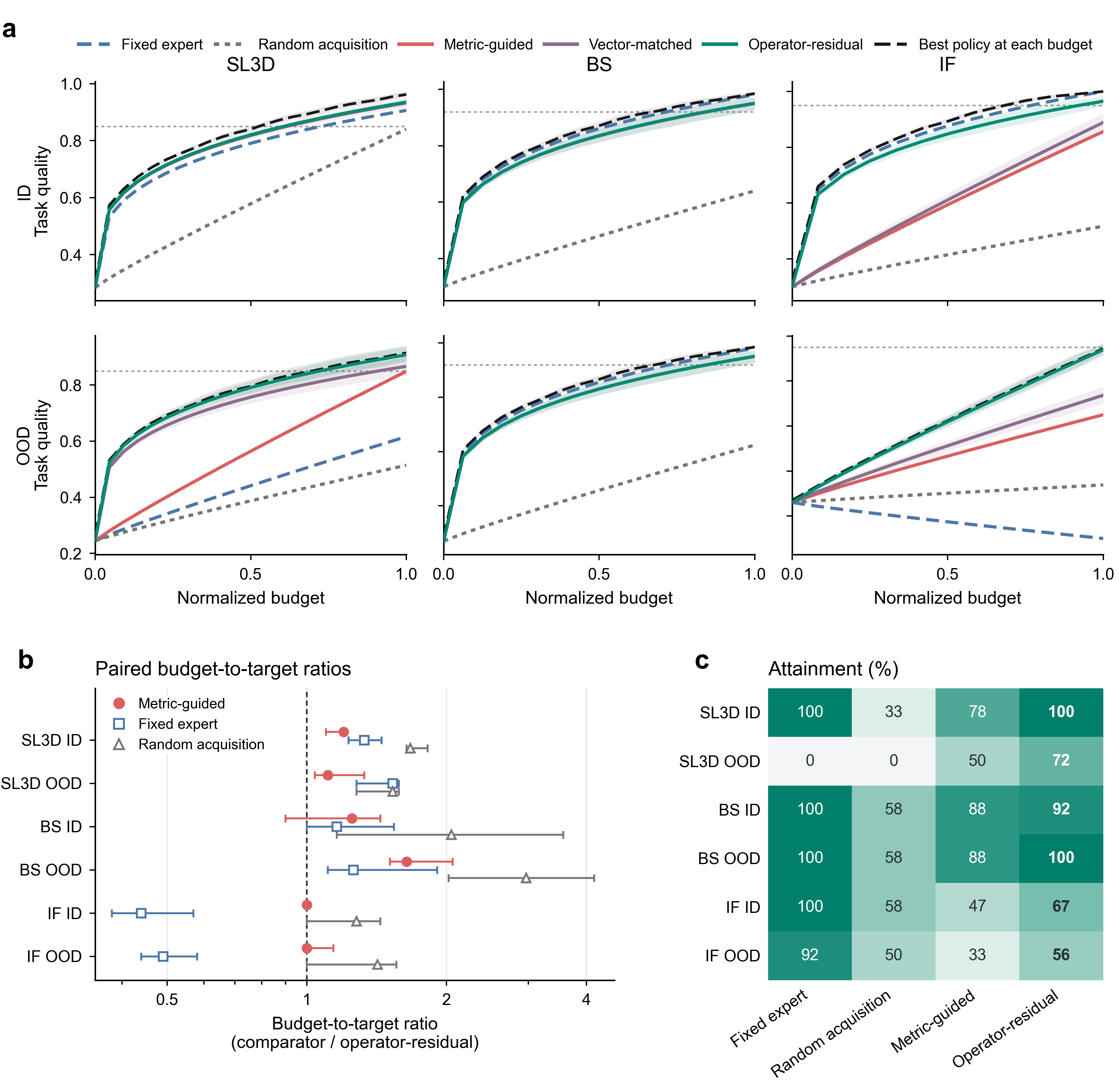}
\caption{\textbf{Operator--residual feedback improves target attainment and stability.}
\textbf{a}, Task performance over normalized budget in the designated display cohort under familiar and prespecified shifted conditions. Curves show problem-weighted means for the four primary reference strategies and operator--residual feedback; dotted horizontal lines mark task targets. Best policy at each budget is the pointwise maximum among the six prespecified strategies and is included as a visual reference rather than an executable strategy because its contributing strategy can change with budget.
\textbf{b}, First-crossing budget ratios in the same display cohort. Each point is the median comparator budget divided by the operator--residual budget, with values above one favoring operator--residual feedback. Bars show 95\% intervals from resampling matched trajectory pairs (Supplementary Table~8).
\textbf{c}, Percentage of trajectories that initially reached the target. Trajectories without observed attainment remain in the denominator. Confirmatory inference required the target to be maintained through the end of evaluation and used 270 independent test problems (Supplementary Tables~10 and 11). Stability after initial attainment in IF is reported in Appendix Fig.~\ref{edfig:if_collapse} and Supplementary Table~9. BS, beam shaping; SL3D, structured-light three-dimensional reconstruction; IF, interferometry; OOD, out of distribution.}
\label{fig:budget}
\end{figure}

Resource use was evaluated with the same stability requirement. Budget was expressed as a percentage of the task-specific cap, and unsuccessful trajectories were assigned the full budget. Operator--residual feedback used 61.7\% of the available budget, compared with 70.8--86.0\% for the four reference strategies. The largest reduction was relative to random acquisition, with lower use also observed relative to the fixed schedule, metric-guided selection and vector-matched feedback. All 24 prespecified task-by-condition contrasts favoured operator--residual feedback after Holm adjustment ($P\leq0.004$; Fig.~\ref{fig:budget} and Supplementary Tables~10 and 11).

Comparisons with stronger algorithmic baselines showed that the budget advantage was condition dependent. Under shifted conditions, operator--residual feedback used 2.6\% less of the full budget than constrained Bayesian optimization and 4.6\% less than the task-specific methods. Under familiar conditions, there was no corresponding advantage over constrained Bayesian optimization, and the task-specific methods used 2.7\% less of the full budget. Terminal $\mathrm{L}_{3}$ was 1.2--2.0\% of the full scale lower with operator--residual feedback across the task-by-condition strata. The observed advantage therefore concerned resource use under distribution shift rather than the highest terminal performance (Supplementary Tables~13--17).

A fixed residual rule tested the same residual channels and candidate operators without a language model. The rule used a prespecified task-specific ranking to select actions. It used 0.8\% less of the full budget under familiar conditions, whereas the language model policy used 2.5\% less under shifted conditions. Across all conditions, the language model policy used 0.9\% less of the full budget (Supplementary Tables~13, 16 and 17). The reversal across conditions indicates that the language model policy adapted operator selection more effectively under distribution shift, whereas the fixed rule remained efficient in familiar settings.

\FloatBarrier

\subsection{Residual feedback guides operator selection}\label{sec:fidelity}

We varied how residual information was represented and then audited the operators selected by the agent. A separate control first compressed all residual violations into a single scalar. The rate of score increases without physical improvement was 39.5\% with score-only feedback, 11.6\% with the aggregate scalar and 2.3\% with the full residual vector. Budget use followed the same ordering (Supplementary Table~18). Thus, adding an aggregate violation scalar recovered part of the missing physical information, while separate residual channels provided an additional benefit.

A second control retained the residual values but varied their interpretation. Restoring stable channel identities, directions, units and operating limits reduced budget use by 8.6\% of the full budget. Adding physical descriptions reduced it by a further 5.3\% (Supplementary Table~16). The agent therefore used not only the magnitude of a residual, but also the physical condition to which it referred.

An offline audit then compared the selected operator with five alternatives executed from the same cloned state. Residual feedback increased selection of the operator with the largest offline improvement in $\mathrm{L}_{3}$ in all three tasks. The largest change occurred in structured-light reconstruction, where the selection rate rose from 2\% to 34\% (Appendix Fig.~\ref{edfig:choice} and Supplementary Table~19). These audit branches were discarded after evaluation and did not alter the live trajectory or expose $\mathrm{L}_{3}$ to the agent.

\begin{figure}[H]
\centering
\mainfig{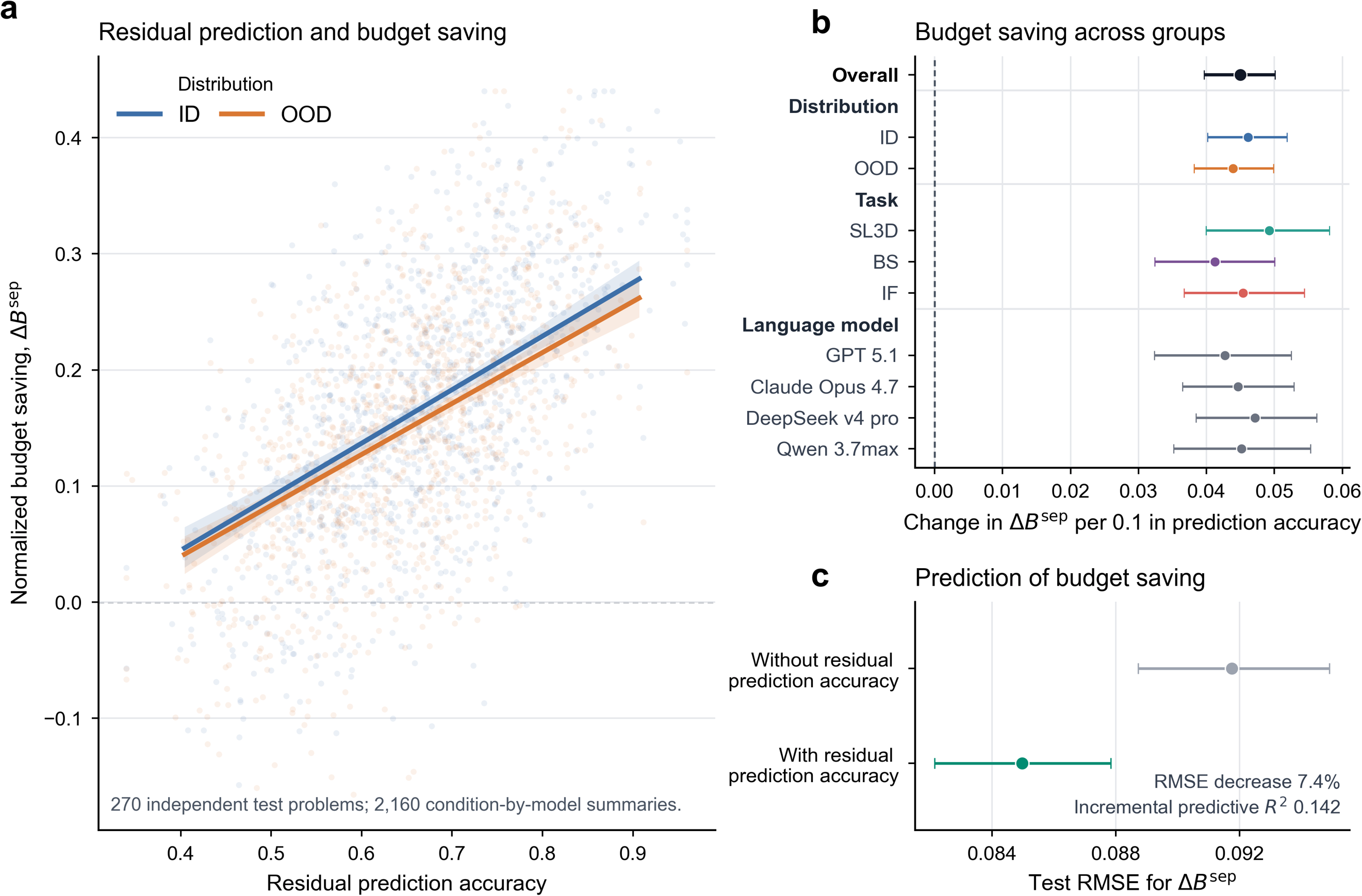}
\caption{\textbf{Residual prediction accuracy is associated with lower budget use.}
\textbf{a}, Residual prediction accuracy in the prespecified seed-1 trajectories versus the budget reduction averaged over separate trajectories using seeds 2--5. Prediction accuracy compares the expected and observed effects of an operator on the residuals. Budget reduction is measured relative to a control that received the same residual values without channel identities or physical descriptions. Lines adjust for initial difficulty, condition, optical task and language model.
\textbf{b}, Adjusted association between residual prediction accuracy and budget reduction per 0.1 increase in prediction accuracy. Points show estimates, bars show 95\% confidence intervals from 5,000 bootstrap resamples of independent problems and the dashed line marks no association.
\textbf{c}, Held-out prediction error after both models were fitted on 90 separate development problems and evaluated on 270 test problems. The second model additionally included residual prediction accuracy and its interaction with condition. Bars show 95\% bootstrap confidence intervals. BS, beam shaping; SL3D, structured-light three-dimensional reconstruction; IF, interferometry.}
\label{fig:fidelity}
\end{figure}

We then tested whether accurate prediction of residual responses was associated with later budget use. Across independent trajectories, a 0.1 increase in prediction accuracy was associated with 4.5\% lower use of the full budget (95\% CI, 4.0--5.0\%; Fig.~\ref{fig:fidelity}a,b and Supplementary Tables~20 and 21). Including prediction accuracy reduced held-out prediction error by 7.4\% (95\% CI, 5.5--9.2\%; Fig.~\ref{fig:fidelity}c). These positive associations support a link between tracking residual responses and operator selection, but the offline analysis does not establish causality. Separate stress tests examined residual omission, monitoring and injected faults (Appendix Fig.~\ref{edfig:controlled_stress}).

\FloatBarrier

\subsection{Physical validation of strategies generated within the OPERA framework}\label{sec:physical_validation}

\begin{figure}[H]
\centering
\fullfig{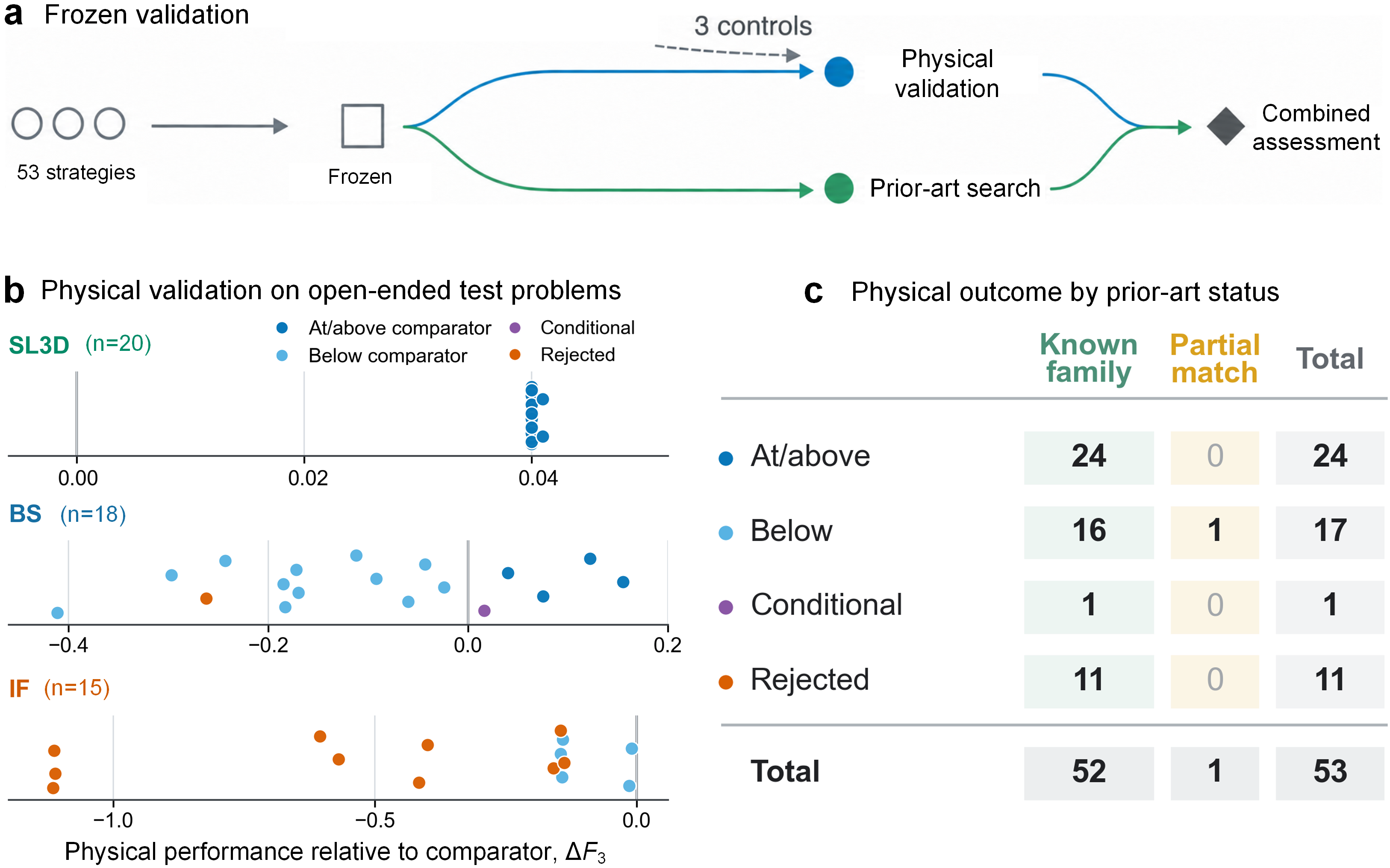}
\caption{\textbf{Physical validation of strategies generated within the OPERA framework.}
\textbf{a}, Frozen validation design. The 60 prespecified attempts comprised 12 from score-only feedback, 24 from operator--residual feedback and 24 from a dedicated synthesis condition. These groups yielded 11, 21 and 21 executable strategies, respectively; all non-executable attempts remained in the generation denominators. Each executable strategy was frozen before independent physical validation and prior-art assessment. Three planted failure controls entered physical validation only.
\textbf{b}, Physical validation on independent BS, SL3D and IF test problems. Each point represents one frozen strategy and shows its task-specific validation result relative to the matched comparator, with positive values favoring the generated strategy. Horizontal scales are task specific. Strategy-level estimates and 95\% confidence intervals are reported in Supplementary Table~24.
\textbf{c}, Physical outcome by prior-art status for the same 53 executable strategies. All three failure controls were rejected and excluded from the literature search. Generation flow is reported in Supplementary Table~22, literature assessments in Supplementary Table~23, strategy-level results in Supplementary Table~24 and task-specific validation definitions in Supplementary Table~25. BS, beam shaping; SL3D, structured-light three-dimensional reconstruction; IF, interferometry.}
\label{fig:physical_validation}
\end{figure}

The common operator interface enabled the agent to generate executable optical strategies. Each generated strategy was frozen before evaluation on independent test problems under the same resource limit as its comparator. Physical validation and prior-art assessment were completed independently before their results were combined (Fig.~\ref{fig:physical_validation}a).

Of 60 prespecified generation attempts, 53 produced executable strategies and 7 did not. Among the executable strategies, 24 met or exceeded their matched comparators, 17 performed below them, one was conditionally accepted and 11 were rejected. All three planted failure controls were rejected (Fig.~\ref{fig:physical_validation}b). In structured-light reconstruction, all 20 executable strategies met or exceeded the comparator, although 17 were variants of the same established coding and phase-reconstruction procedure.

The prior-art assessment assigned 52 executable strategies to established algorithm families and one to a partial match; none was unmatched. Eleven strategies associated with recognized families nevertheless failed physical validation. Executability, physical performance and prior-art status were therefore distinct outcomes, and generation within the common interface did not by itself establish novelty (Fig.~\ref{fig:physical_validation}c and Supplementary Tables~22--25).

\FloatBarrier

\subsection{Transfer of digital-twin protocols to optical hardware}\label{sec:hardware}

Protocols selected in the digital twins were frozen before transfer to the three optical instruments, with no online updates from hardware residuals. This design tested the executability of the selected actions and acquisition sequences under physical measurement conditions.

\begin{figure}[p]
\centering
\includegraphics[width=\textwidth,height=\dimexpr\textheight-52pt\relax,keepaspectratio]{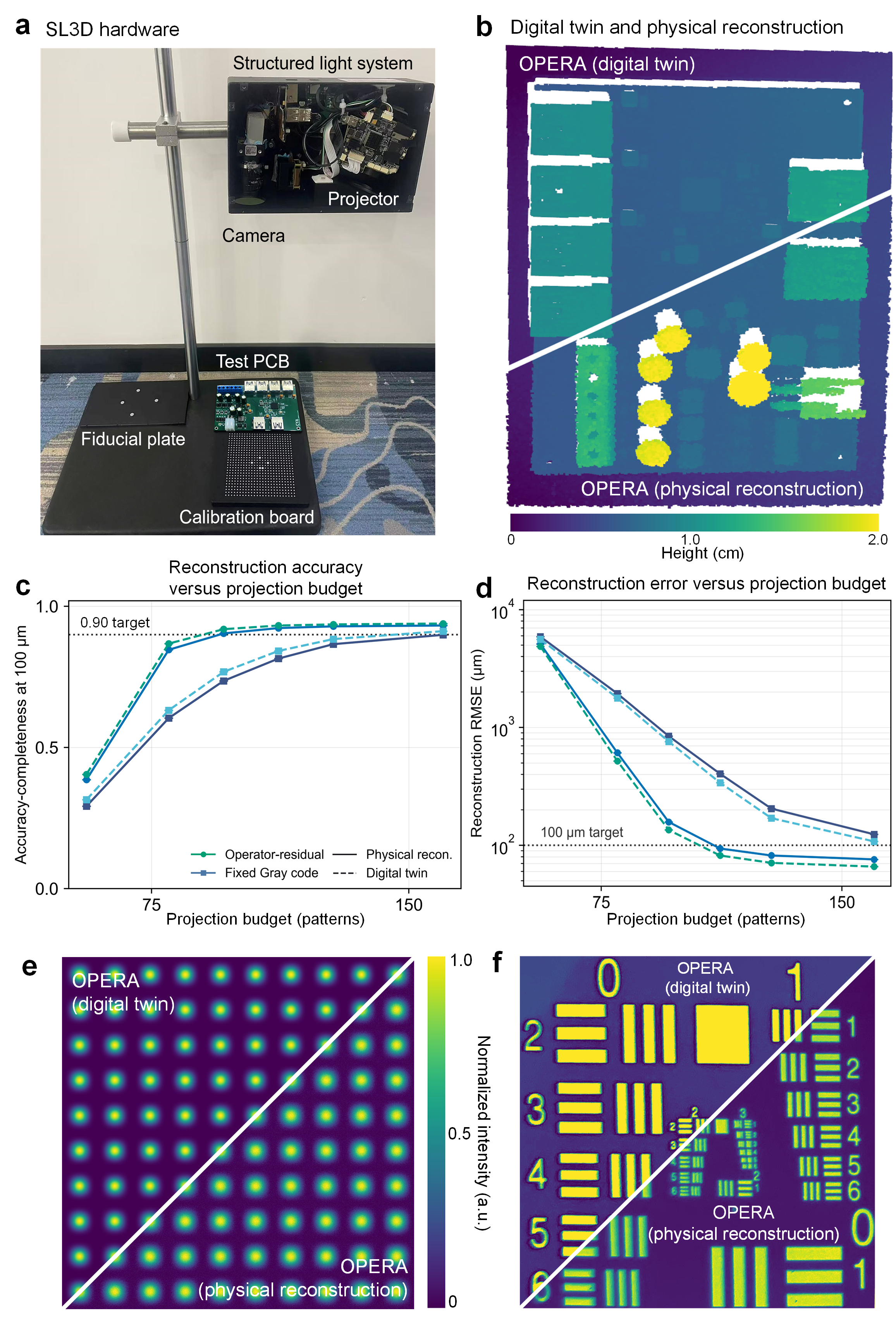}
\caption{\textbf{Transfer of digital-twin protocols to optical hardware.}}
\label{fig:hardware}
\end{figure}

Structured-light reconstruction provided the repeated budget-to-target comparison. Three hardware configurations were each evaluated in six paired repeats. The transferred protocol used a lower mean fraction of the projection budget than fixed Gray code in every configuration; in the principal configuration, the corresponding means were 60.0\% and 95.8\% of the available budget. It also reached the reconstruction target with fewer projections in both the digital twin and the physical system (Fig.~\ref{fig:hardware}a--d and Appendix Fig.~\ref{edfig:sl3d_dense}).

Beam shaping and interferometry tested physical reconstruction after transfer. In beam shaping, the selected phase mask produced a spot array with a mean full-width at half-maximum of 0.67 Airy units in the physical reconstruction, compared with 0.59 in the digital twin and the target value of 0.50 (Fig.~\ref{fig:hardware}e and Appendix Fig.~\ref{edfig:bs_reconstruction}). In interferometry, phase was inferred from the acquired intensity sequence rather than measured directly. The transferred acquisition sequence produced a registered intensity reconstruction with a normalized cross-correlation of 0.96 relative to the reference state (Fig.~\ref{fig:hardware}f and Appendix Fig.~\ref{edfig:if_profile}). Structured-light reconstruction therefore supplied repeated evidence for lower budget use across three configurations (Appendix Fig.~\ref{edfig:sl3d_replication}), while beam shaping and interferometry extended physical transfer to modulation and interferometric acquisition. Supplementary Videos~1--3 show the corresponding digital-twin execution trajectories.

\begin{figure}[!t]
\noindent{\small \textbf{\figurename~\thefigure\ (continued).} \textbf{a}, Physical SL3D system. The structured-light projector and camera were registered with a calibration board and fiducial plate before imaging the printed-circuit-board specimen.
\textbf{b}, Digital-twin and physical reconstructions of the same centimeter-scale specimen, separated by the diagonal line and displayed on a common height scale.
\textbf{c}, Reconstruction accuracy at 100~\textmu m versus projection budget for the operator--residual and fixed Gray-code protocols in the principal configuration. One budget unit comprises one projected pattern and the corresponding camera acquisition. Solid curves show physical reconstruction, dashed curves show the digital twin and the dotted line marks the hardware-specific target of 0.90. This target is distinct from the simulator target $\mathrm{L}_{3,\mathrm{sim}}=0.85$ used in Fig.~\ref{fig:budget}. Protocols were frozen before transfer and received no online updates from hardware residuals.
\textbf{d}, Reconstruction root-mean-square error against the independently acquired dense reference over the same projection-budget axis. Colours and line styles follow \textbf{c}, and the dotted line marks a 100~\textmu m reference level. Repeat-level observations and uncertainty are reported in Appendix Fig.~\ref{edfig:sl3d_replication}.
\textbf{e}, Normalized Fourier-plane intensity for the BS phase mask selected within the OPERA framework in the digital twin and physical reconstruction, separated by the diagonal line.
\textbf{f}, Normalized intensity for the IF protocol selected within the OPERA framework in the digital twin and physical reconstruction of a reflective 1951 USAF resolution target. The quantitative IF metrics compare registered intensity patterns. Quantitative BS and IF audits are reported in Appendix Figs.~\ref{edfig:bs_reconstruction} and \ref{edfig:if_profile}; repeat-level SL3D results across all three configurations are reported in Appendix Fig.~\ref{edfig:sl3d_replication}. See also Supplementary Videos~1--3. BS, beam shaping; SL3D, structured-light three-dimensional reconstruction; IF, interferometry.}
\end{figure}
\FloatBarrier

\section{Discussion}\label{sec:discussion}

Autonomous optical experiments require both a clear set of executable actions and feedback that reveals the physical effects of those actions. OPERA provides these elements through operators and residuals. Operators define changes to measurement, modulation and reconstruction, while residuals report departures from specified physical conditions. Across beam shaping, structured-light reconstruction and interferometry, this action--feedback interface reduced score increases without physical improvement and increased the probability of reaching and maintaining task targets. The central contribution is therefore an interface that guides autonomous decisions using executable optical actions and measurable physical evidence.

The controls clarify how the structure and interpretation of residual feedback affected physical validity. Compressing all residual violations into one scalar recovered part of the benefit over score-only feedback, while retaining separate channels produced a further improvement. Stable identities, operating limits and physical descriptions helped the agent interpret those channels. In the offline audit, residual feedback also increased selection of the operator with the largest physical improvement. Residual prediction accuracy was positively associated with lower budget use, although this observational analysis does not establish causality. More broadly, feedback design determines which failures an autonomous system can detect \cite{Christiano2017,HadfieldMenell2017}, and multiple physical checks can expose faults hidden from a single output \cite{Chow1984}.

Target attainment shows whether the target was maintained, whereas budget use shows how many resources were required. Operator--residual feedback increased the probability that task targets were reached and maintained, although later loss of target performance was not eliminated. Its budget advantage over stronger algorithmic baselines was concentrated under shifted conditions and was accompanied by slightly lower terminal $\mathrm{L}_{3}$. The fixed residual rule used less budget in familiar settings, whereas the language model policy used less under distribution shift. The residual design supplied the physical information, while the language model policy translated that information into operator selection, combination and generation through a common interface. The language model policy contributed most under shifted conditions, where operator selection had to adapt beyond familiar task settings.

The common operator interface produced 53 executable strategies from 60 prespecified attempts, and 24 met or exceeded their matched comparators. Independent validation also showed why execution, physical performance and prior-art status must be assessed separately. Most executable strategies implemented established algorithm families, and some failed physical validation despite that match. OPERA therefore supports systematic generation and evaluation of optical strategies through a shared interface, while novelty remains a separate outcome assessed against prior art.

The hardware experiments showed that protocols selected in digital twins remained executable across three physical instruments with distinct optical operations and measurements. In structured-light reconstruction, the transferred protocol retained a projection-budget advantage across three hardware configurations. Beam shaping and interferometry confirmed the physical execution and evaluation of the selected phase mask and acquisition sequence. Freezing the protocols before transfer separated hardware performance from further online adaptation and allowed direct comparison with the digital-twin results. The same operator--residual interface can also be used in hardware loops that return residuals from live measurements to the agent. Recent systems have shown that language-model agents can interact with laboratory instruments during operation \cite{Vriza2026Instruments,Chen2026XRay}, and related closed-loop control has been demonstrated in optical trapping \cite{Selin2026SmartTrap}. OPERA contributes a structured physical feedback interface that can support such operation.

The three implementations also illustrate how the framework can be extended. The decision loop is separated from the task-specific operators and residuals. Operators can be defined to measure an object, modify an optical field or adjust an instrument state, and they can be combined to connect quantitative measurement, structural reconstruction and active control. Residuals provide the corresponding physical checks at each stage. Extending OPERA to a new instrument therefore requires defining and validating its available operations and observable physical checks, while the overall decision loop remains unchanged. This modularity does not remove the need for instrument knowledge. It places that knowledge in explicit operators and residuals that can be inspected, validated and expanded. By making experimental actions and physical evidence explicit, OPERA provides a route towards autonomous experiments guided by measurable physical outcomes.

\section{Conclusion}\label{sec:conclusion}
OPERA turns two requirements of autonomous experimentation, executable actions and physically grounded feedback, into a single typed interface between a language-model agent and an optical instrument. Across three optical tasks, four language models and three physical instruments, operator--residual feedback suppressed score increases without physical improvement, raised the probability of reaching and maintaining task targets and reduced experimental budgets, with the remaining advantage over strong algorithmic baselines concentrated under distribution shift. Because the decision loop is separated from the task-specific operators and residuals, extending the framework to a new instrument requires defining and validating its available operations and observable physical checks. This provides a route towards autonomous experiments held to measurable physical evidence.

\paragraph{Data availability}
The decision-level dataset, recorded traces, strategy-level physical-validation outcomes and source data underlying all figures are publicly available on Zenodo at DOI: \href{https://doi.org/10.5281/zenodo.21504254}{\nolinkurl{10.5281/zenodo.21504254}}.

\paragraph{Code availability}
The OPERA framework implementation, agent gateway, counterfactual-audit scripts and analysis code are publicly available at \url{https://github.com/ningxu1995/OPERA}.

\paragraph{Supplementary information}
Supplementary Methods, Supplementary Tables 1--27 and Supplementary Videos 1--3 are provided as ancillary files with this arXiv submission.

\paragraph{Acknowledgements}
This work is partially supported by Tsinghua University (Department of Computer Science and Technology)--Sinopec Joint Research Center for Artificial Intelligence.

\paragraph{Author contributions}
N.X. conceptualized the study. X.Z. and H.W. developed the methodology. N.X., X.Z., H.W. and Z.L. conducted the investigation. N.X. and X.Z. performed the formal analysis and validation. N.X., X.Z., F.Z. and H.W. curated the data and prepared the visualizations. Z.L. acquired funding. Z.L. and H.N. supervised the work. All authors contributed to writing the original draft.

\paragraph{Competing interests}
The authors declare no competing interests.

\bibliographystyle{unsrtnat}
\bibliography{sn-bibliography}

\appendix
\clearpage
\section{Extended results figures}\label{app:figs}
\renewcommand{\thefigure}{A\arabic{figure}}
\setcounter{figure}{0}
\noindent This appendix reproduces the extended data figures that accompany the journal version of this manuscript. Cross-references of the form Appendix Fig.~A$n$ in the main text point here.

\begin{figure}[H]
\centering
\fullfig{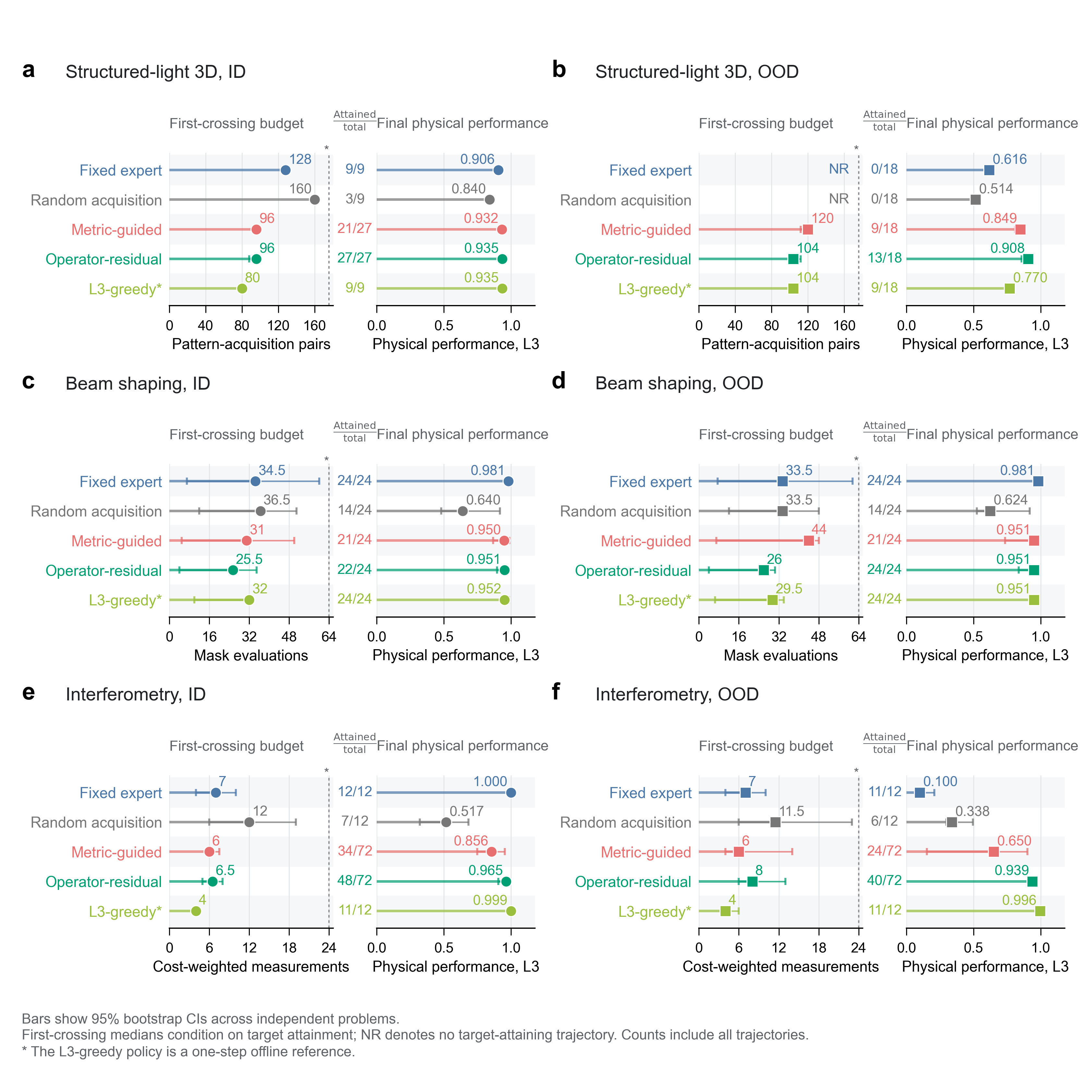}
\caption{\textbf{Target crossing and final physical performance across tasks.}
For each task and condition, the panels report the median first-crossing budget among target-attaining trajectories, the number of trajectories that attained the target and final physical performance, $\mathrm{L}_{3}$. Horizontal bars show 95\% bootstrap confidence intervals across independent problems for first-crossing budget and final $\mathrm{L}_{3}$. NR indicates that no trajectory attained the target; dashed vertical lines mark the task-specific budget cap. Fixed expert, random acquisition, metric-guided selection and operator--residual feedback are executable strategies. The $\mathrm{L}_{3}$-greedy reference instead uses withheld $\mathrm{L}_{3}$ to select the candidate with the largest immediate $\Delta\mathrm{L}_{3}$. It is myopic and non-executable, and it does not define an upper bound. Best policy at each budget in Fig.~\ref{fig:budget}a is a different visual reference calculated after execution from six budget-matched strategies. Budget units are mask evaluations for BS, projected-pattern and camera-acquisition pairs for SL3D, and cost-weighted measurements for IF. Source values are reported in Supplementary Tables~7 and 9. BS, beam shaping; SL3D, structured-light three-dimensional reconstruction; IF, interferometry; ID, familiar condition; OOD, prespecified shifted condition.}
\label{edfig:budget_trajectories}
\end{figure}
\FloatBarrier
\clearpage

\begin{figure}[H]
\centering
\fullfig{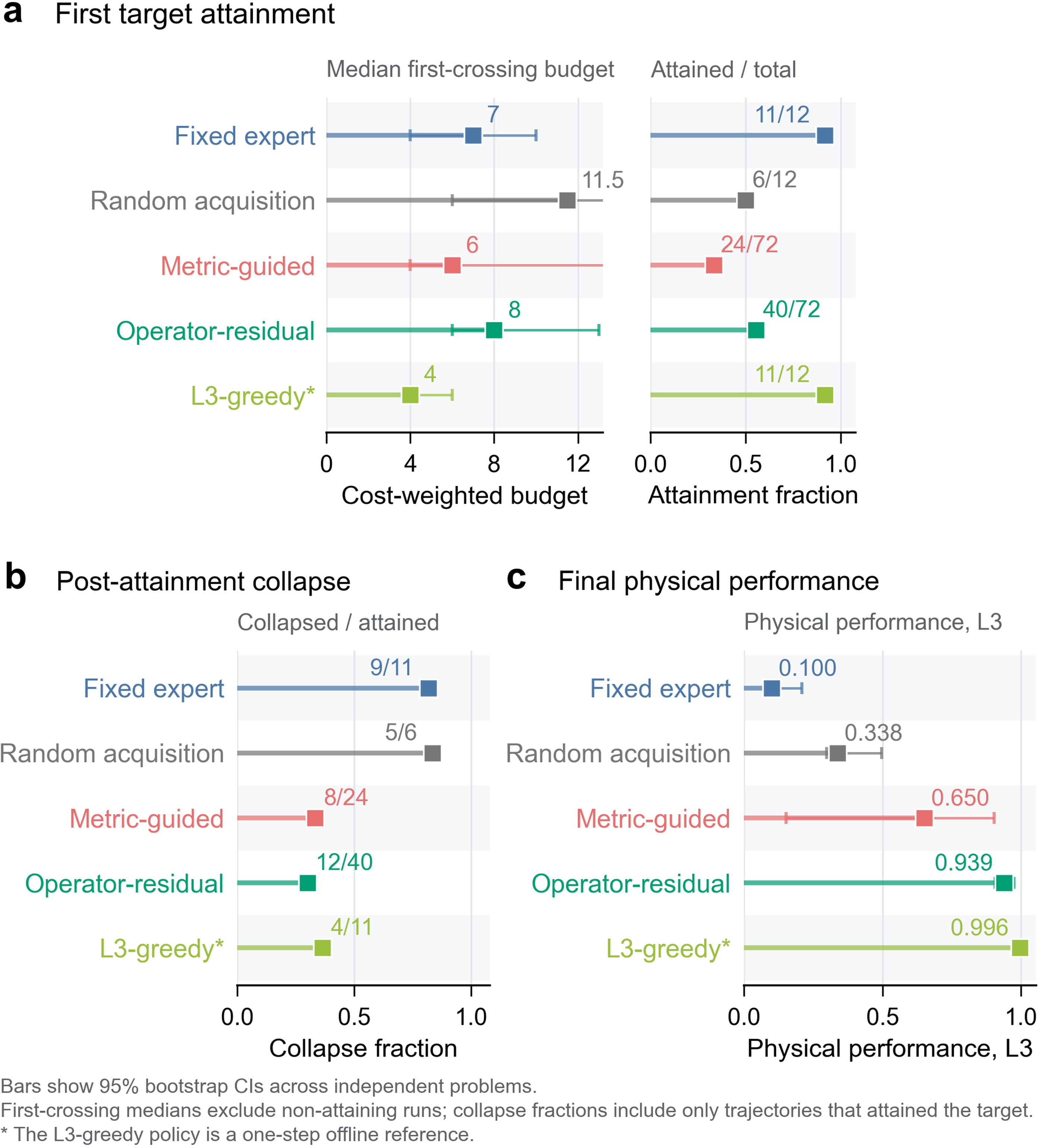}
\caption{\textbf{Initial target attainment does not always persist in interferometry.}
\textbf{a}, Median first-crossing budget among target-attaining trajectories and the number attaining the target under the shifted condition. Trajectories without attainment are excluded from the median but retained in the attainment fraction.
\textbf{b}, Fraction of target-attaining trajectories that later fell below the target before the budget cap.
\textbf{c}, Final physical performance, $\mathrm{L}_{3}$, across all submitted trajectories. Horizontal bars in \textbf{a} and \textbf{c} show 95\% bootstrap confidence intervals across independent problems. Together, the panels separate initial crossing, subsequent stability and final performance. The non-executable $\mathrm{L}_{3}$-greedy reference is defined in Appendix Fig.~\ref{edfig:budget_trajectories}. Counts and endpoint definitions are reported in Supplementary Table~9.}
\label{edfig:if_collapse}
\end{figure}
\clearpage

\begin{figure}[H]
\centering
\includegraphics[width=\textwidth,height=0.80\textheight,keepaspectratio]{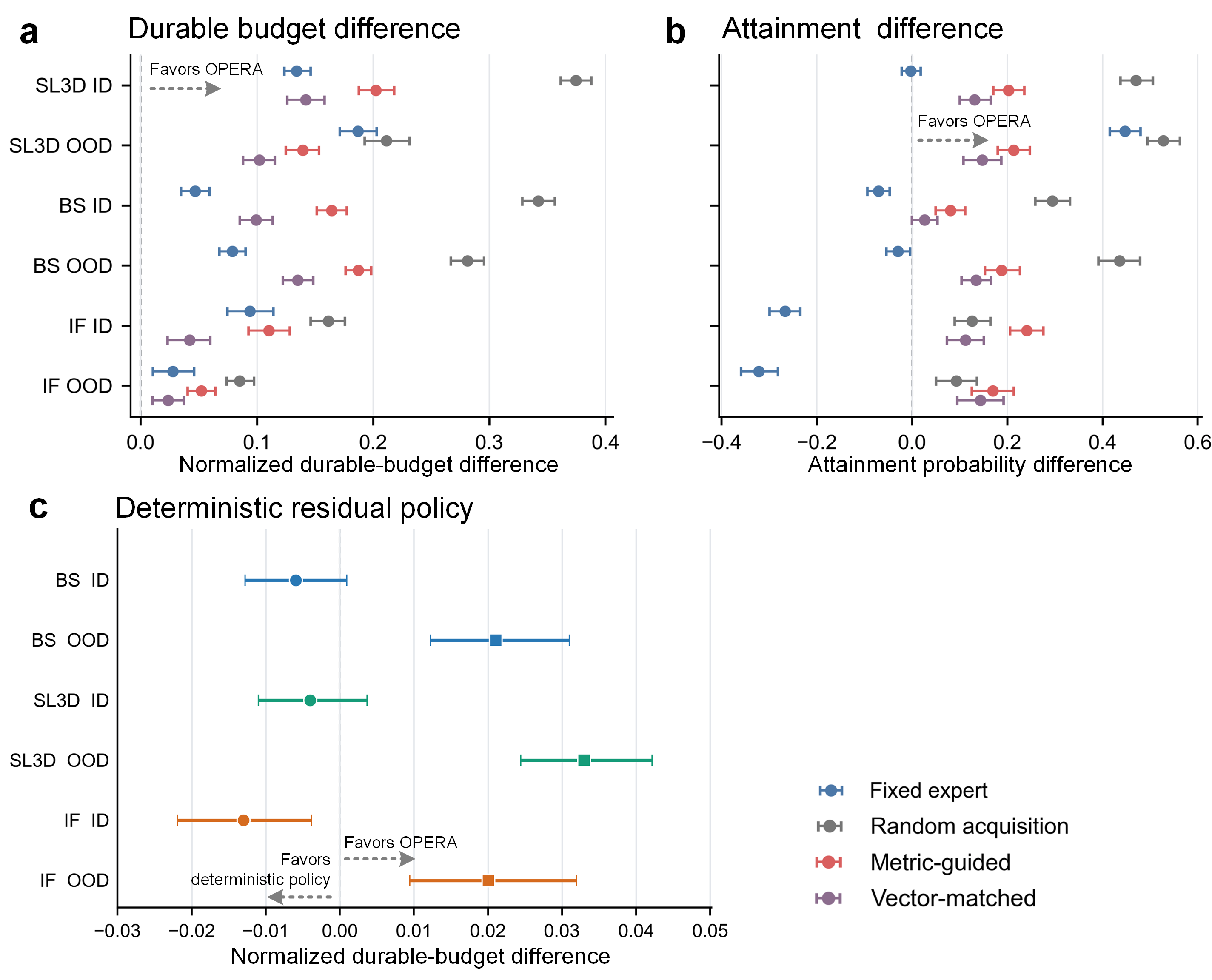}
\caption{\textbf{Confirmatory budget and attainment comparisons across independent test problems.}
\textbf{a}, Normalized durable-budget difference, defined as $b_{\mathrm{comparator}}-b_{\mathrm{operator\mbox{-}residual}}$. Positive values indicate lower durable budget under operator--residual feedback.
\textbf{b}, Attainment-probability difference, defined as $p_{\mathrm{operator\mbox{-}residual}}-p_{\mathrm{comparator}}$. Positive values indicate higher attainment under operator--residual feedback.
\textbf{c}, Paired durable-budget comparison with the fixed residual rule. The plotted value is $b_{\mathrm{fixed}}-b_{\mathrm{operator\mbox{-}residual}}$; positive values indicate lower budget under operator--residual feedback. The fixed rule receives the same $\mathrm{L}_{1}$ channels, candidate operators and budget, but applies a prespecified ranking without a language model or access to $\mathrm{L}_{3}$. Repeated records were first averaged within each of 270 independent problems. Points show mean paired effects, bars show 95\% confidence intervals from 5,000 bootstrap resamples of problem roots and the dashed line marks no difference. These confirmatory estimates are separate from the display cohort in Fig.~\ref{fig:budget}. Full comparator estimates are reported in Supplementary Table~16; the primary strategy comparisons are reported in Supplementary Tables~10 and 11. BO, Bayesian optimization; BS, beam shaping; SL3D, structured-light three-dimensional reconstruction; IF, interferometry; ID, familiar condition; OOD, prespecified shifted condition.}
\label{edfig:confirmatory_budget}
\end{figure}
\clearpage

\begin{figure}[H]
\centering
\includegraphics[width=0.90\linewidth]{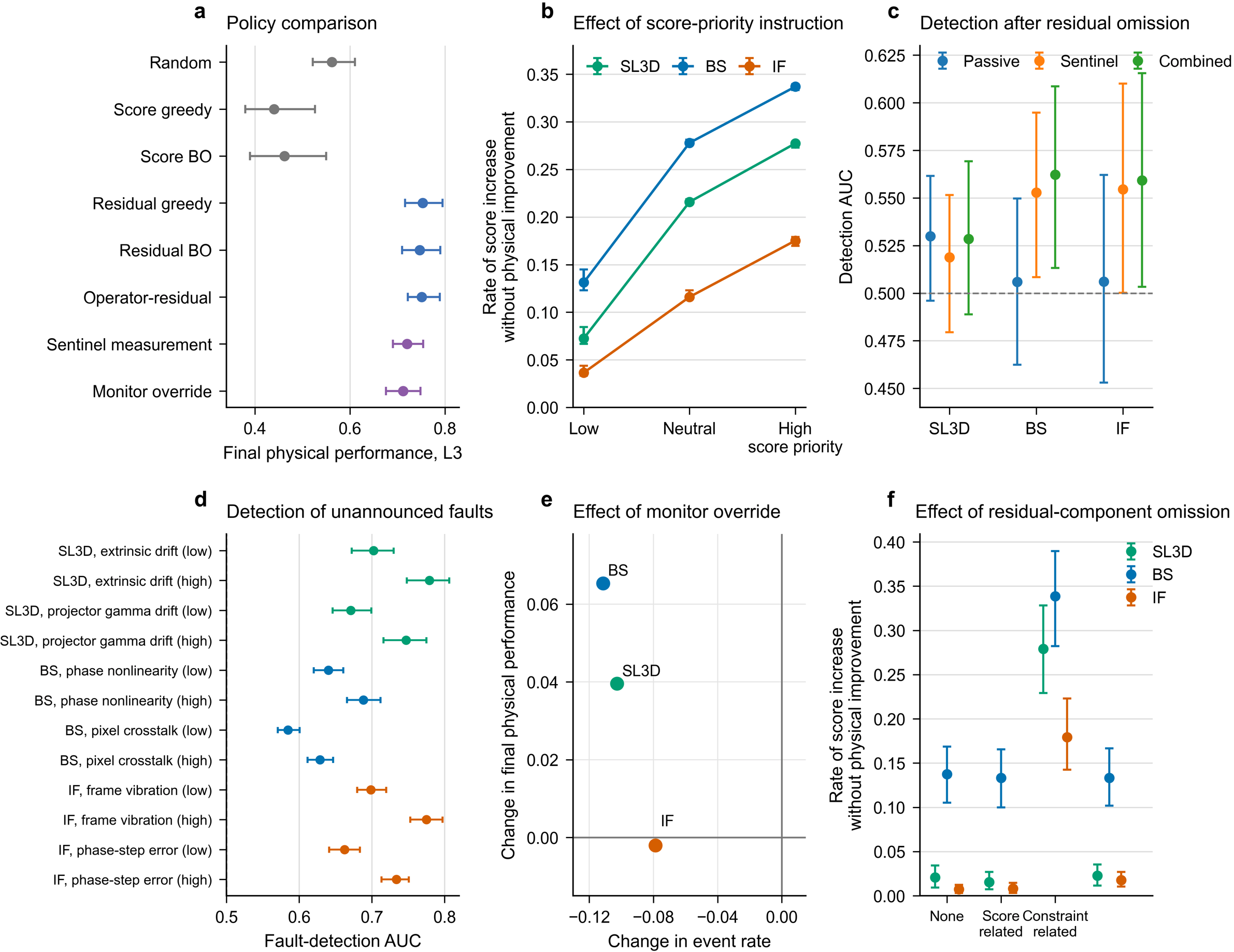}
\caption{\textbf{Residual feedback increases selection of the operator with the largest physical improvement.}
Points show the fraction of audited decisions for which the selected operator produced the largest offline $\Delta\mathrm{L}_{3}$ among six candidates. Open circles denote Score only and filled squares denote Score + residual. Pale lines connect paired estimates within each task. Horizontal bars show the observed ranges across four BS, three SL3D and six IF problem blocks; they are not confidence intervals. Uniform selection gives a reference rate of $1/6$. The audit used cloned simulator states, and no branch altered the live trajectory or its budget. Source values, audited-decision counts and exact block-level sign-flip tests are reported in Supplementary Table~19.}
\label{edfig:choice}
\end{figure}
\clearpage

\begin{figure}[H]
\centering
\includegraphics[width=\textwidth,height=0.82\textheight,keepaspectratio]{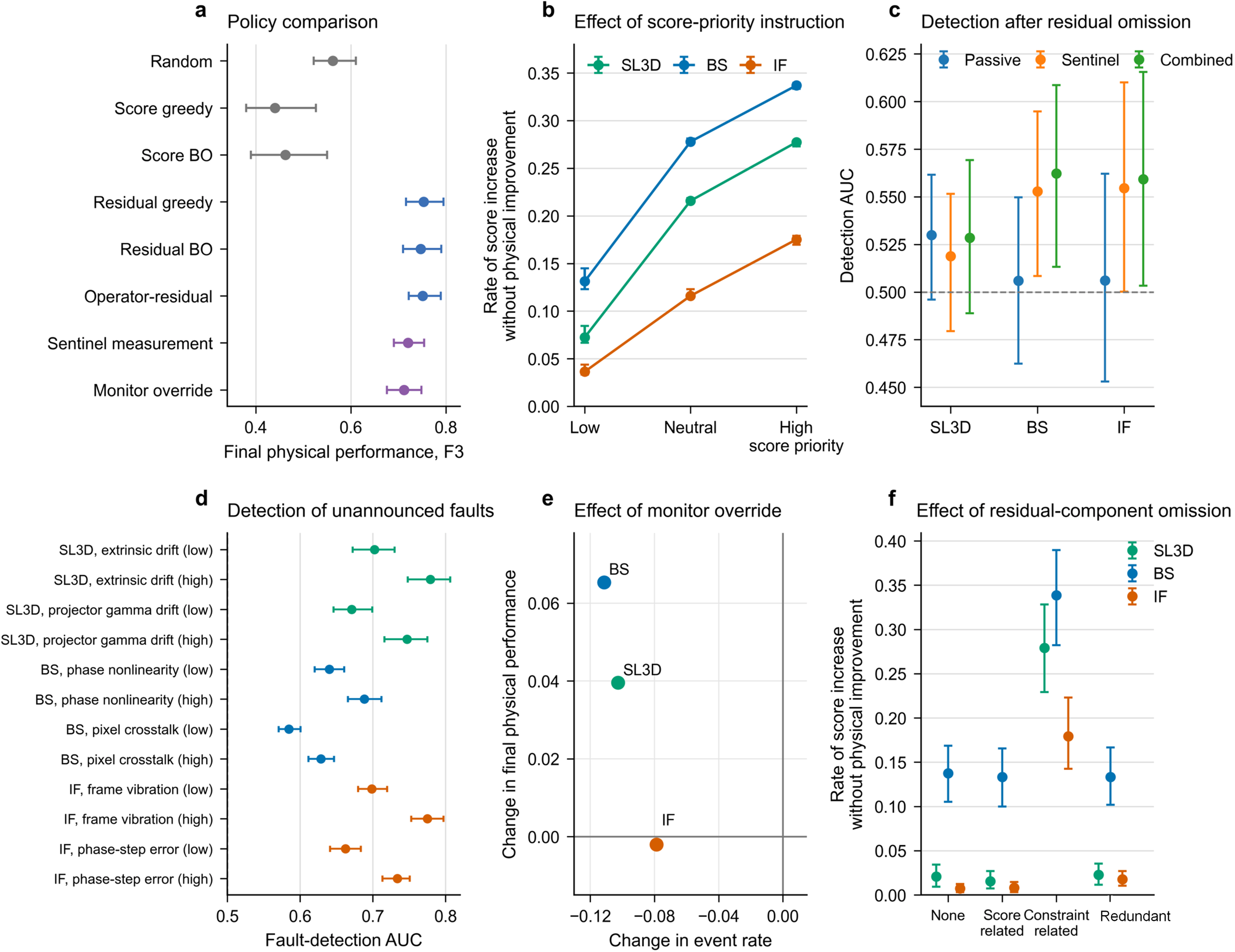}
\caption{\textbf{Controlled stress tests isolate residual omissions, injected faults and monitoring responses.}}
\label{edfig:controlled_stress}
\end{figure}
\begin{figure}[H]
\noindent{\small \textbf{\figurename~\thefigure\ (continued).} These simulator stress tests were analysed separately from the primary three-task and hardware experiments. Thirty clean familiar-condition problems were used only to set monitor thresholds; a different set of 120 problems was used for evaluation. Stress-test instructions and override rules are defined in Supplementary Methods.
\textbf{a}, Final physical performance, $\mathrm{L}_{3}$, with no injected fault, all residuals present and a Neutral instruction. Points show means and bars span the six task-by-condition combinations. Each combination contains 20 independent evaluation problems and two repeated trajectories.
\textbf{b}, Rate of score increase without physical improvement after residual components were omitted and the instruction placed low, neutral or high priority on score increase. Bars span five equivalent phrasings of the instruction after pooling familiar and shifted problems within each task.
\textbf{c}, Detection of score increases without physical improvement after residual omission. Passive monitor uses the existing residual stream, Sentinel measurement adds a budget-consuming measurement and Combined monitor uses both. Bars show 95\% confidence intervals from resampling evaluation problems; the dashed line marks chance performance.
\textbf{d}, Detection of six unannounced faults at low and high strength by the combined monitor. Bars show 95\% CIs obtained by resampling evaluation problems. H, high; L, low.
\textbf{e}, Change produced by a monitor override relative to the same policy without an override. A lower event rate and higher final physical performance define the favourable upper-left quadrant. Monitoring and override costs are charged within the shared 12-unit cap.
\textbf{f}, Rate of score increase without physical improvement after omitting the score-related, constraint-related or redundant residual component. Bars show 95\% CIs obtained by resampling evaluation problems. AUC, area under the receiver operating characteristic curve; BO, Bayesian optimization; BS, beam shaping; SL3D, structured-light three-dimensional reconstruction; IF, interferometry.}
\end{figure}
\clearpage

\begin{figure}[H]
\centering
\includegraphics[width=\textwidth,height=\dimexpr\textheight-52pt\relax,keepaspectratio]{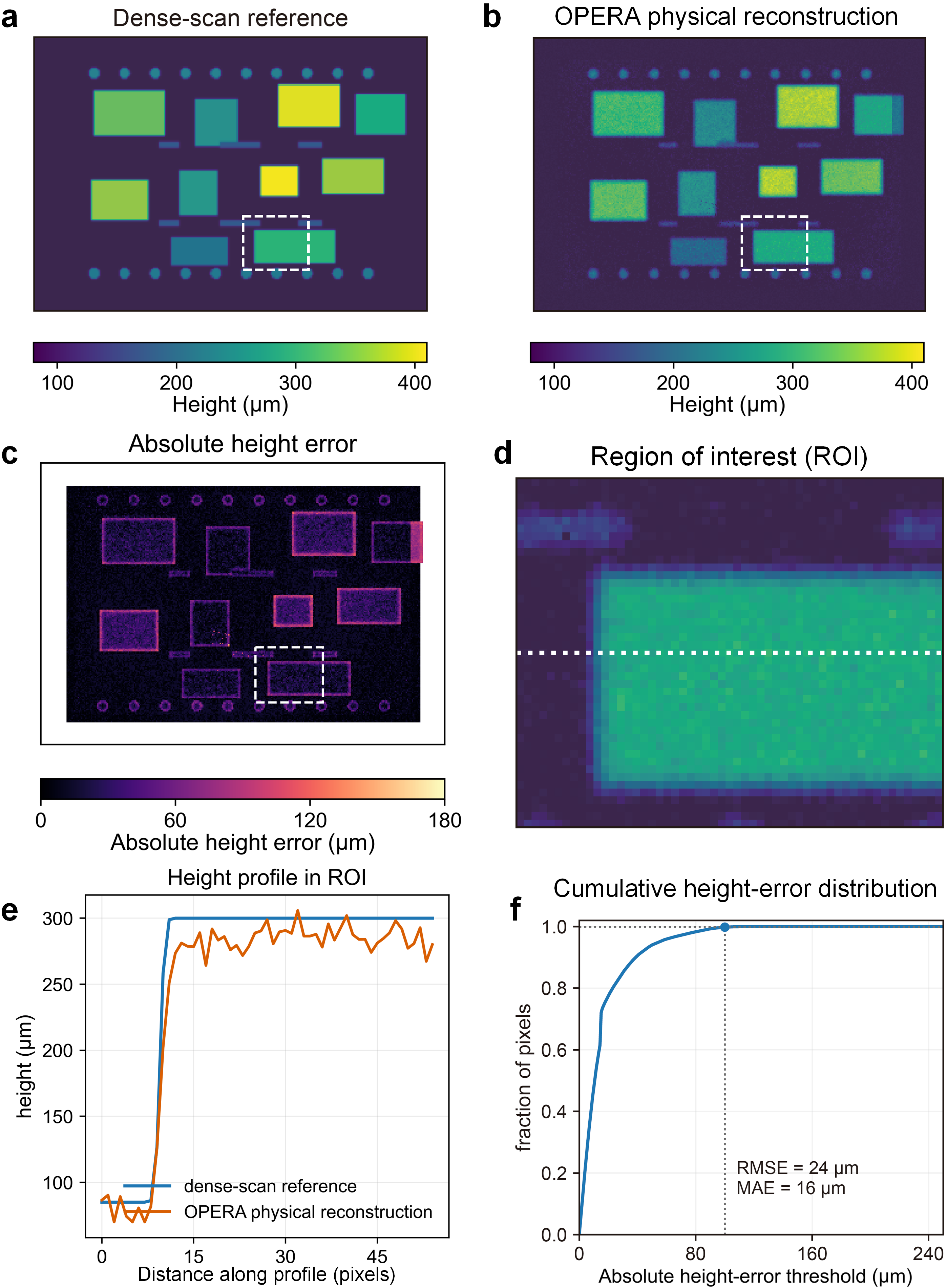}
\caption{\textbf{A transferred structured-light protocol reconstructs micrometer-scale relief.}}
\label{edfig:sl3d_dense}
\end{figure}
\begin{figure}[H]
\noindent{\small \textbf{\figurename~\thefigure\ (continued).} \textbf{a}, Dense-scan reference height map of the micrometer-scale printed-circuit-board specimen.
\textbf{b}, Physical reconstruction obtained by executing the projection protocol selected within the OPERA framework on the same specimen.
\textbf{c}, Absolute height error after registration to the dense-scan reference; the dashed box marks the region of interest used only for the local profile analysis.
\textbf{d}, Region of interest used for the profile comparison.
\textbf{e}, Height profiles extracted from the dense-scan reference and the physical reconstruction along the line shown in \textbf{d}.
\textbf{f}, Empirical cumulative distribution of absolute height error over all valid registered pixels in the prespecified full reconstruction area. The reported root-mean-square error and mean absolute error use the same pixels; the region in \textbf{d,e} is not used for these global summaries. The vertical line marks the 100~\textmu m tolerance. This specimen differs from the centimeter-scale board shown in Fig.~\ref{fig:hardware}b and tests the same projection protocol at a complementary axial scale. The dense-scan reference was used only for offline physical evaluation and was never available to the decision loop.}
\end{figure}
\clearpage

\begin{figure}[H]
\centering
\includegraphics[width=\textwidth,height=\dimexpr\textheight-52pt\relax,keepaspectratio]{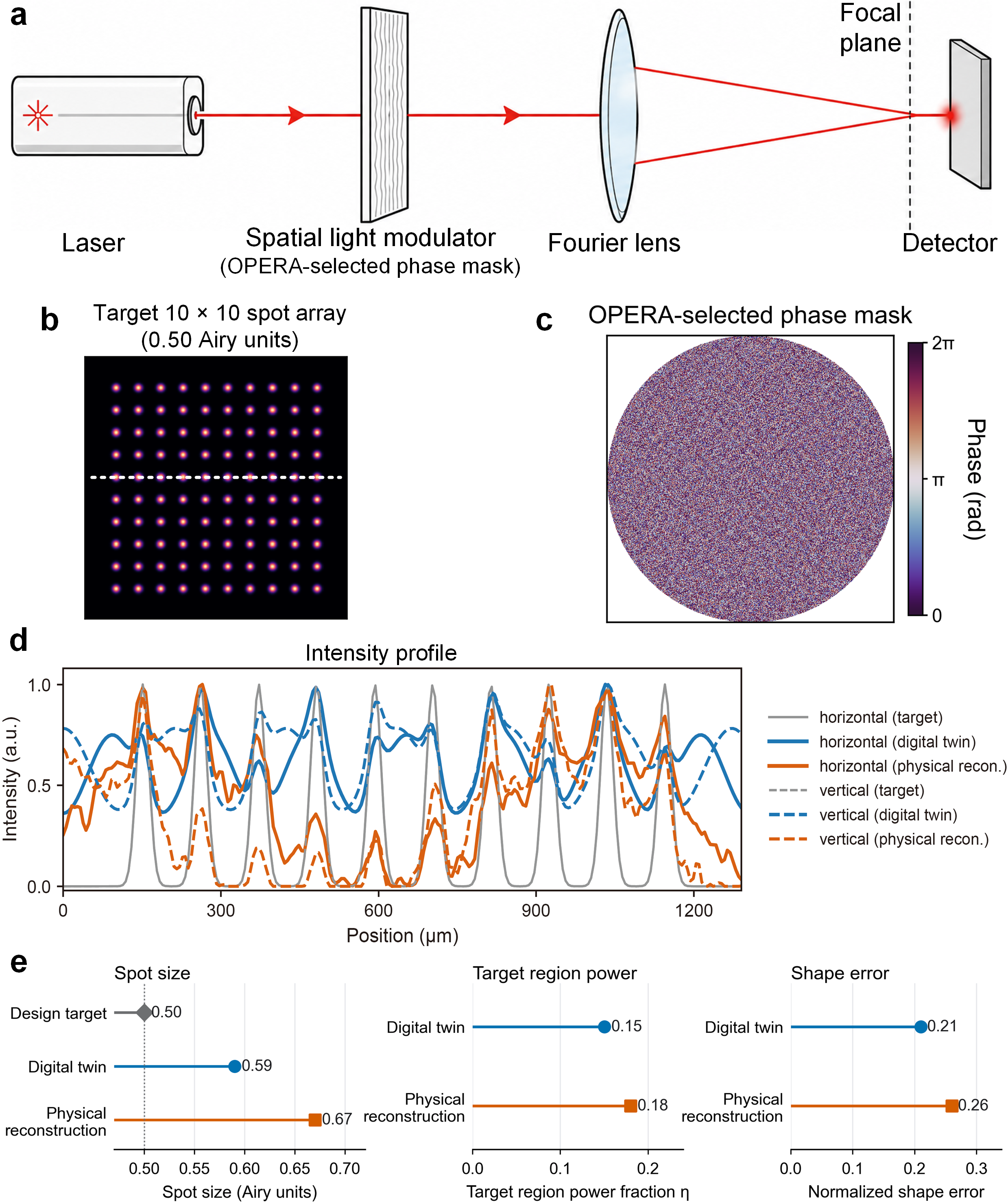}
\caption{\textbf{Physical reconstruction from a transferred beam-shaping mask.}}
\label{edfig:bs_reconstruction}
\end{figure}
\begin{figure}[H]
\noindent{\small \textbf{\figurename~\thefigure\ (continued).} \textbf{a}, Optical layout. A He--Ne laser illuminated a spatial light modulator carrying the phase mask selected within the OPERA framework; a 200~mm Fourier lens formed the shaped field at the detector.
\textbf{b}, Target $10 \times 10$ spot array with a target spot size of 0.50 Airy units; the white dashed line marks the profile used in \textbf{d}.
\textbf{c}, Wrapped phase mask selected within the OPERA framework and displayed on the spatial light modulator.
\textbf{d}, Horizontal and vertical intensity profiles from the target, digital twin and physical reconstruction at the positions marked in \textbf{b}. Each profile is normalized to unit peak intensity.
\textbf{e}, Spot size in Airy units, target-region power fraction $\eta$ and normalized shape error for the displayed mask. The spot sizes were 0.59 Airy units in the digital twin and 0.67 Airy units in the physical reconstruction. The corresponding values were $\eta=0.15$ and shape error 0.21 in the digital twin, and $\eta=0.18$ and shape error 0.26 in the physical reconstruction. The displayed values are point summaries for the principal mask loading; additional measurements are supplied with the source data.}
\end{figure}
\clearpage

\begin{figure}[H]
\centering
\includegraphics[width=\textwidth,height=\dimexpr\textheight-52pt\relax,keepaspectratio]{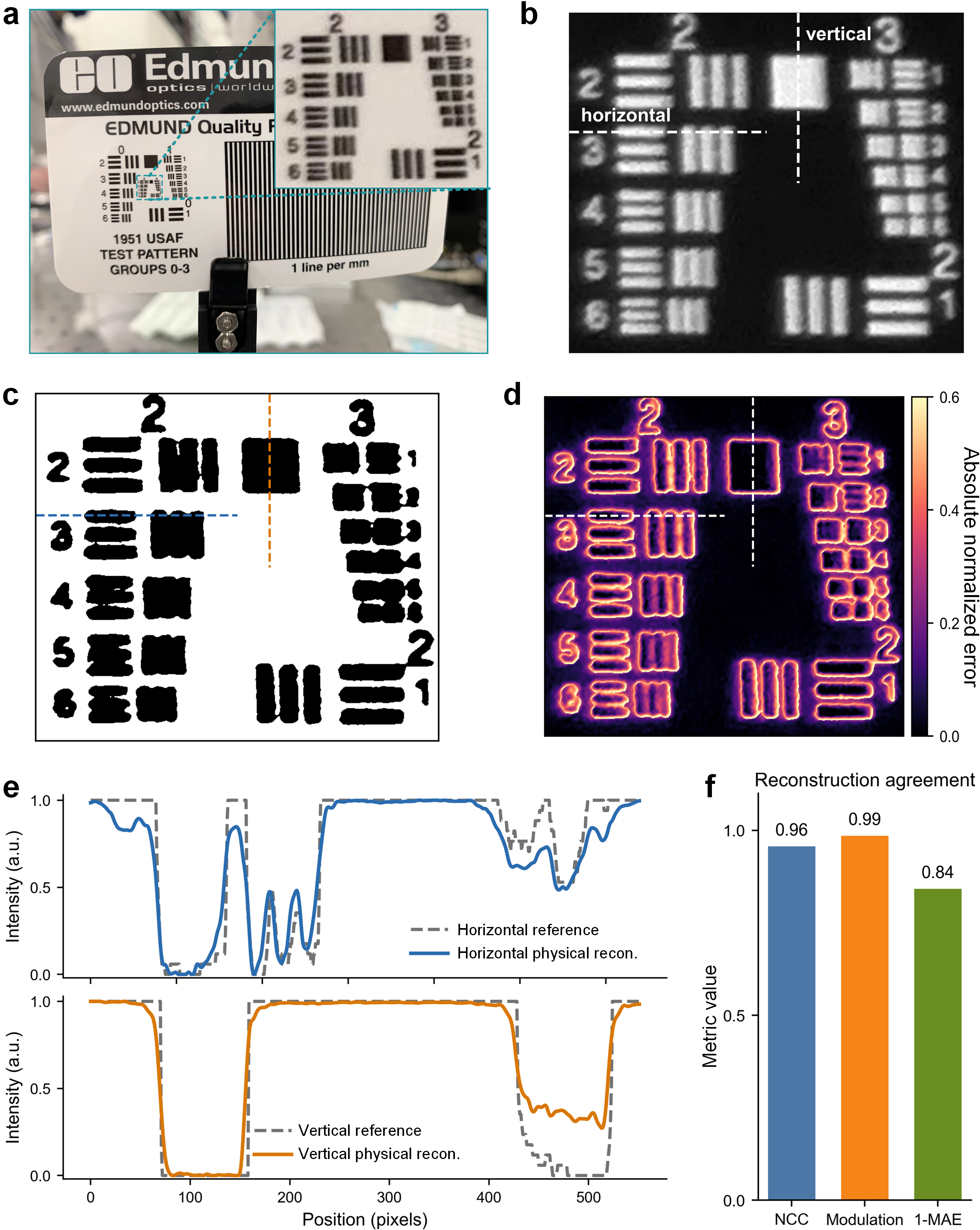}
\caption{\textbf{A transferred interferometry protocol reconstructs a reflective resolution target.}}
\label{edfig:if_profile}
\end{figure}
\begin{figure}[H]
\noindent{\small \textbf{\figurename~\thefigure\ (continued).} \textbf{a}, Photograph of the reflective 1951 USAF resolution target; the marked region indicates the local evaluation area.
\textbf{b}, Measured intensity in the selected local region, normalized to unit peak intensity. Dashed lines mark the horizontal and vertical profiles used in \textbf{e}.
\textbf{c}, Binary reference template derived from the selected subregion and used for the local comparison.
\textbf{d}, Absolute normalized error between the physical reconstruction and the local reference template.
\textbf{e}, Horizontal and vertical intensity profiles. Gray dashed curves denote the reference and solid colored curves denote the physical reconstruction.
\textbf{f}, Reconstruction agreement in the registered local region. Normalized cross-correlation (NCC) measures spatial agreement; modulation is the robust contrast computed from the 5th and 95th intensity percentiles; and $1-\mathrm{MAE}$ is one minus the mean absolute error after unit-range normalization. The displayed metrics are point summaries for the acquisition shown; additional hardware configurations are supplied with the source data.}
\end{figure}
\clearpage

\begin{figure}[H]
\centering
\includegraphics[width=\textwidth,height=0.82\textheight,keepaspectratio]{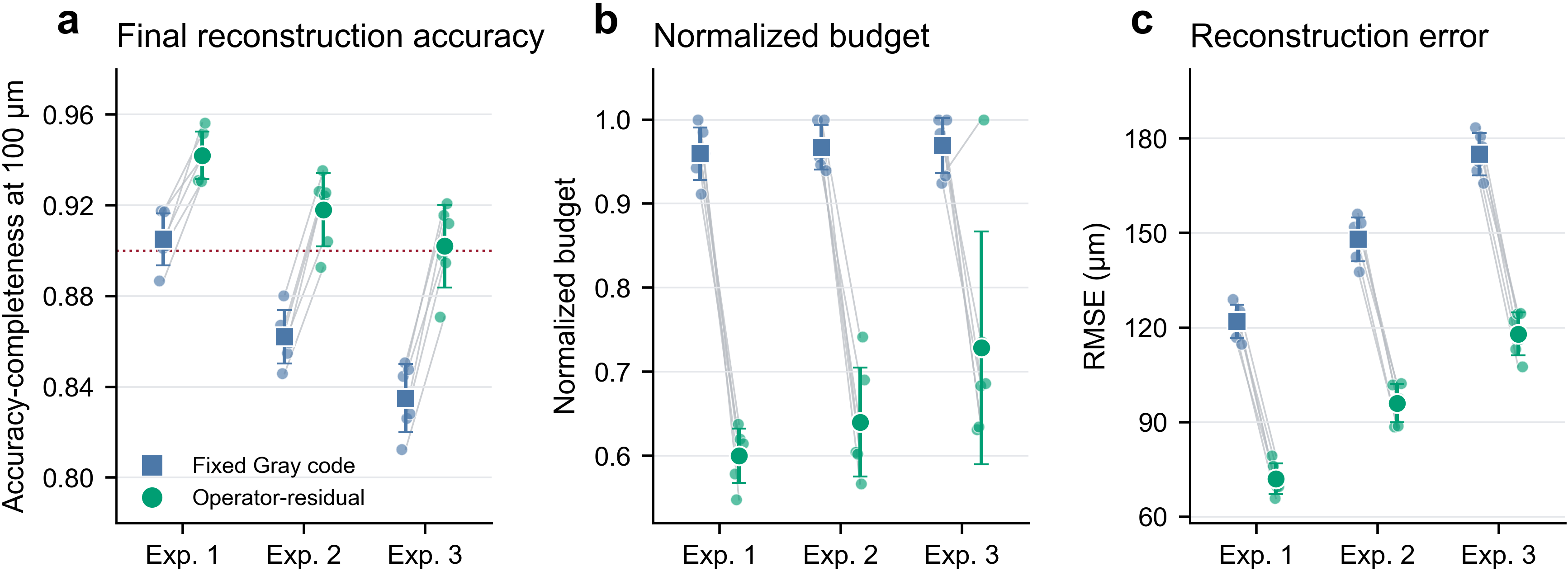}
\caption{\textbf{Repeated structured-light transfer across three hardware configurations.}
Each panel compares the operator--residual and fixed Gray-code protocols. \textbf{a}, Final reconstruction accuracy, measured as accuracy-completeness at 100~\textmu m.
\textbf{b}, Normalized first-crossing budget to the hardware-specific accuracy-completeness target; lower values indicate fewer projected-pattern and camera-acquisition pairs.
\textbf{c}, Reconstruction root-mean-square error against an independently acquired dense reference; lower values indicate better reconstruction.
Small points show individual paired technical repeats, gray lines connect the same repeat across protocols, large symbols show means and bars show standard deviations ($n=6$ paired technical repeats per experiment; 18 paired repeat units in total). Open symbols in \textbf{b} denote trajectories for which no target crossing was observed; these trajectories were assigned the prespecified budget cap. Experiment 1 is the principal configuration shown in Fig.~\ref{fig:hardware}; Experiments 2 and 3 are additional hardware configurations. The micrometer-scale specimen in Appendix Fig.~\ref{edfig:sl3d_dense} is separate and does not enter these repeat sets. Paired-repeat bootstrap estimates are supplied with the source data. SL3D, structured-light three-dimensional reconstruction.}
\label{edfig:sl3d_replication}
\end{figure}
\clearpage

\end{document}